\documentclass[11pt]{article}
\usepackage{sibarticle}
\usepackage{lmodern}
\usepackage{url}            
\usepackage{nicefrac}       
\usepackage{microtype}      
\usepackage{amssymb,graphicx}
\usepackage{float,amsmath}
\usepackage{amsfonts,epsfig}
\usepackage{tabularx}
\usepackage{xspace}
\usepackage{comment}

\usepackage{subcaption}
\usepackage{color}
\usepackage{algorithm}
\usepackage[noend]{algpseudocode}
\algblock{ParFor}{EndParFor}
\algnewcommand\algorithmicparfor{\textbf{parfor}}
\algnewcommand\algorithmicpardo{\textbf{do}}
\algnewcommand\algorithmicendparfor{\textbf{end\ parfor}}
\algrenewtext{ParFor}[1]{\algorithmicparfor\ #1\ \algorithmicpardo}
\algrenewtext{EndParFor}{\algorithmicendparfor}
\newcommand{\mocrt}{\texttt{M-OCRT}\xspace}
\newcommand{\eocrt}{\texttt{E-OCRT}\xspace}
\newcommand{\epocrt}{\texttt{EP-OCRT}\xspace}
\newcommand{\mrf}{\texttt{M-RF}\xspace}
\newcommand{\erf}{\texttt{E-RF}\xspace}
\newcommand{\prf}{\texttt{EP-RF}\xspace}
\newcommand{\rf}{\texttt{RF}\xspace}
\newcommand{\skl}{\texttt{SKLEARN}\xspace}
\newcommand{\ocrt}{\texttt{OCRT}\xspace}
\newcommand{\floor}[1]{\left\lfloor #1 \right\rfloor}

\title{Output-Constrained Regression Trees} 
\ShortTitle{OCRT}
\ShortAuthors{Tunç, Özese, Birbil, Maragno, Caserta, Baydoğan}

\NumberOfAuthors{6}
\FirstAuthor{Hüseyin Tunç}
\FirstAuthorAddress{Social Sciences University of Ankara, Turkey}
\SecondAuthor{Doğanay Özese}
\SecondAuthorAddress{Uber Amsterdam, The Netherlands}
\ThirdAuthor{Ş. İlker Birbil}
\ThirdAuthorAddress{University of Amsterdam, The Netherlands}
\FourthAuthor{Donato Maragno}
\FourthAuthorAddress{Amazon, Germany}
\FifthAuthor{Marco Caserta}
\FifthAuthorAddress{Amazon, Luxembourg}
\SixthAuthor{Mustafa Baydoğan}
\SixthAuthorAddress{Boğaziçi University, Turkey}


\allowdisplaybreaks

\begin{document}
\setuptodonotes{backgroundcolor=red!10, linecolor=red,
  bordercolor=red, size=\tiny, tickmarkheight=0.1cm}

\newcommand\hlb[3][]{ \todo[inline,caption={emptytext},
  size=\normalsize, backgroundcolor=yellow!70, bordercolor=yellow!70, noshadow, #1]{
    \begin{minipage}{
        \textwidth-4pt}#2
    \end{minipage}}
  \todo{\begin{spacing}{0.5}#3\end{spacing}}}

\newcommand{\hlc}[2]{\hl{#1}\todo{\begin{spacing}{0.5}#2\end{spacing}}}

\newcommand{\sitodo}[2][]{\todo[caption={#2}, #1]{
    \begin{spacing}{0.5}#2\end{spacing}}}

\newcommand{\intodo}[2][]{\todo[inline, noshadow, caption={emptytext}, #1]{
    \begin{spacing}{1.0}\normalsize{#2}\end{spacing}}}

\newcommand{\sib}[1]{\textcolor{violet}{#1}}
\newcommand{\tsib}[1]{\begin{tcolorbox}\textcolor{violet}{#1}\end{tcolorbox}}

\newcounter{sibcmntcounter}
\setcounter{sibcmntcounter}{1}
\long\def\symbolfootnote[#1]#2{\begingroup
  \def\thefootnote{\fnsymbol{footnote}}\footnote[#1]{#2}\endgroup}
\newcommand{\sibcmnt}[1]{{\small\textbf{
      \textcolor{violet}{(C.\arabic{sibcmntcounter})}}
    \let\thefootnote\relax\footnotetext{\textcolor{violet}
        {\scriptsize(C.\arabic{sibcmntcounter})~#1}}}
  \addtocounter{sibcmntcounter}{1}}

\newcommand{\red}[1]{\textcolor{red}{#1}}
\newcommand{\blue}[1]{\textcolor{blue}{#1}}
\newcommand{\magenta}[1]{\textcolor{magenta}{#1}}
\newcommand{\bm}[1]{\mbox{\boldmath{$#1$}}}
\newcommand{\rb}[1]{\raisebox{-1.5ex}[0cm][0cm]{#1}}
\newcommand{\HRule}{\noindent\rule{\linewidth}{0.5mm}}
\newcommand{\dsum}{\displaystyle\sum}
\newcommand{\veps}{\varepsilon}

\newcommand{\CA}{\mathcal{A}}
\newcommand{\CB}{\mathcal{B}}
\newcommand{\CC}{\mathcal{C}}
\newcommand{\CD}{\mathcal{D}}
\newcommand{\CG}{\mathcal{G}}
\newcommand{\CI}{\mathcal{I}}
\newcommand{\CJ}{\mathcal{J}}
\newcommand{\CK}{\mathcal{K}}
\newcommand{\CL}{\mathcal{L}}
\newcommand{\CN}{\mathcal{N}}
\newcommand{\CP}{\mathcal{P}}
\newcommand{\CS}{\mathcal{S}}
\newcommand{\CT}{\mathcal{T}}
\newcommand{\CX}{\mathcal{X}}
\newcommand{\YY}{\mathbb{Y}}
\newcommand{\ZZ}{\mathbb{Z}}
\newcommand{\RR}{\mathbb{R}}
\newcommand{\NN}{\mathbb{N}}
\newcommand{\II}{\mathbb{1}}

\newcommand{\va}{\bm{a}}
\newcommand{\vb}{\bm{b}}
\newcommand{\vc}{\bm{c}}
\newcommand{\vd}{\bm{d}}
\newcommand{\ve}{\bm{e}}
\newcommand{\vf}{\bm{f}}
\newcommand{\vg}{\bm{g}}
\newcommand{\vh}{\bm{h}}
\newcommand{\vp}{\bm{p}}
\newcommand{\vr}{\bm{r}}
\newcommand{\vq}{\bm{q}}
\newcommand{\vt}{\bm{t}}
\newcommand{\vu}{\bm{u}}
\newcommand{\vv}{\bm{v}}
\newcommand{\vw}{\bm{w}}
\newcommand{\vx}{\bm{x}}
\newcommand{\vy}{\bm{y}}
\newcommand{\vz}{\bm{z}}
\newcommand{\zv}{\bm{0}}
\newcommand{\ov}{\bm{1}}

\newcommand{\vveps}{\bm{\veps}}
\newcommand{\veta}{\bm{\eta}}
\newcommand{\vxi}{\bm{\xi}}
\newcommand{\valpha}{\bm{\alpha}}
\newcommand{\vbeta}{\bm{\beta}}
\newcommand{\vgamma}{\bm{\gamma}}
\newcommand{\vtheta}{\bm{\theta}}
\newcommand{\vlambda}{\bm{\lambda}}
\newcommand{\vnu}{\bm{\nu}}
\newcommand{\vpi}{\bm{\pi}}
\newcommand{\vtau}{\bm{\tau}}

\newcommand{\mA}{\bm{A}}
\newcommand{\mB}{\bm{B}}
\newcommand{\mC}{\bm{C}}
\newcommand{\mD}{\bm{D}}
\newcommand{\mE}{\bm{E}}
\newcommand{\mF}{\bm{F}}
\newcommand{\mG}{\bm{G}}
\newcommand{\mH}{\bm{H}}
\newcommand{\mI}{\bm{I}}
\newcommand{\mL}{\bm{L}}
\newcommand{\mM}{\bm{M}}
\newcommand{\mP}{\bm{P}}
\newcommand{\mQ}{\bm{Q}}
\newcommand{\mR}{\bm{R}}
\newcommand{\mS}{\bm{S}}
\newcommand{\mU}{\bm{U}}
\newcommand{\mV}{\bm{V}}
\newcommand{\mX}{\bm{X}}

\newcommand{\tr}{^{\intercal}}
\newcommand{\ntr}{^{-\intercal}}
\newcommand{\inv}{^{-1}}

\newcommand{\gr}{\mbox{graph}}
\newcommand{\ra}{\rightarrow}
\newcommand{\la}{\leftarrow}
\newcommand{\Ra}{\Rightarrow}
\newcommand{\rra}{\rightrightarrows}
\newcommand{\ptr}{\marginpar{$\Leftarrow$}}

\newcommand{\pfxi}{\frac{\partial f(\vx)}{\partial x_i}}
\newcommand{\pfx}{\partial f(\vx)}
\newcommand{\pf}{\partial f}
\newcommand{\pxi}{\partial x_i}
\newcommand{\px}{\partial x}

\newcommand{\nfx}{\nabla f(\vx)}
\newcommand{\eps}{\epsilon}
\newcommand{\eg}{\textit{e.g.}}
\newcommand{\ie}{\textit{i.e.}}

\newcommand{\vsp}{\vspace{4mm}}
\newcommand{\vspp}{\vspace{8mm}}
\newcommand{\vsppp}{\vspace{12mm}}

\newcommand{\hsp}{\hspace{4mm}}
\newcommand{\hspp}{\hspace{8mm}}
\newcommand{\hsppp}{\hspace{12mm}}

\newcommand{\pr}[1]{\mathbb{P}\left(#1\right)}
\newcommand{\ex}[1]{\mathbb{E}\left[#1\right]}
\newcommand{\variance}[1]{\mbox{Var}\left(#1\right)}
\newcommand{\covar}[1]{\mbox{Cov}\left(#1\right)}
\newcommand{\C}[2]{\left(\begin{array}{c} #1 \\ #2 \end{array}\right)}

\newcommand{\maximize}{\mbox{maximize\hspace{4mm} }}
\newcommand{\minimize}{\mbox{minimize\hspace{4mm} }}
\newcommand{\subto}{\mbox{subject to\hspace{4mm}}}

\newenvironment{sibitemize}{
  \renewcommand{\labelitemi}{$\diamond$}
  \begin{itemize}
    \setlength{\parskip}{0mm}}
  {\end{itemize}}

\newcommand{\propnum}[2]{\vspace{3mm}
  \noindent {\sc Proposition #1}{\it #2} \vspace{3mm}}
\newcommand{\lemnum}[2]{\vspace{3mm}
  \noindent {\sc Lemma #1}{\it #2} \vspace{3mm}}
\newcommand{\thmnum}[2]{\vspace{3mm}
  \noindent {\sc Theorem #1}{\it #2} \vspace{3mm}}

\maketitle

\begin{abstract}
  Incorporating domain-specific constraints into machine learning models is essential for generating predictions that are both accurate and feasible in real-world applications. This paper introduces new methods for training Output-Constrained Regression Trees (\ocrt), addressing the limitations of traditional decision trees in constrained multi-target regression tasks. We propose three approaches: \mocrt, which uses split-based mixed integer programming to enforce constraints; \eocrt, which employs an exhaustive search for optimal splits and solves constrained prediction problems at each decision node; and \epocrt, which applies post-hoc constrained optimization to tree predictions. To illustrate their potential uses in ensemble learning, we also introduce a random forest framework working under convex feasible sets. We validate the proposed methods through a computational study both on synthetic and industry-driven hierarchical time series datasets. Our results demonstrate that imposing constraints on decision tree training results in accurate and feasible predictions.
\end{abstract}

\section{Introduction.}
\label{sec:intro}

The performances of machine learning (ML) models often benefit from incorporating domain knowledge. While most ML research focuses on single-output prediction, where the goal is to estimate a single target variable, many real-world applications require predicting multiple interdependent outputs simultaneously. In such cases, treating each output separately can overlook dependencies, leading to suboptimal results.  While existing methodologies—such as shape-constrained models \citep[e.g.,][]{AitSahalia2003}—successfully regulate input-output relationships by enforcing properties like convexity or monotonicity, a fundamental gap remains: they do not ensure that the predicted outputs strictly satisfy domain-specific constraints reflecting their interdependency. Furthermore, as shape-constrained frameworks do not guarantee the operational feasibility of test instances, they fall short in practical applications where output constraints are strict prerequisites.  This oversight is significant because, in many practical applications, such output constraints are critical for ensuring the feasibility of predictions. For example, predicted values may need to sum to a predefined total, maintain hierarchical consistency, or adhere to logical dependencies. Without satisfying these constraints, the predictions themselves may not be usable for decision-makers regardless of their accuracy.

The present work stems from the following important observation: Although many ML algorithms can be constrained during training to respect these feasibility restrictions for train data, they all fail to guarantee constraint satisfaction for unseen test data.  This is particularly critical when the nature of the problem requires integer predictions or must satisfy sophisticated logical and mathematical dependencies, which inherently transform even a simple convex range into a non-convex feasible set. For instance, in logistics and inventory management, predictions may involve discrete lot sizes or the number of vehicles required for a shipment, where outputs must be integers. Furthermore, many applications involve logical ``either-or'' dependencies, such as minimum order quantities where a prediction for a product must be either zero or above a specific threshold, or prohibited operating zones in energy systems where certain output ranges must be avoided for safety. These constraints result in disjoint, non-convex feasible regions that standard regression models—which typically predict continuous averages—cannot navigate.  However, decision trees (DTs), when equipped with appropriate training algorithms, have the unique potential among modern ML methods to address this gap. By exploring this potential, we propose new DT training algorithms, which guarantee predictions satisfying given constraints among multiple output variables. DTs are arguably the most frequently used ML methods in practice due to their versatility in both classification and regression tasks across various domains, including finance, healthcare, marketing, and engineering. DT training algorithms can easily handle both categorical and numerical data, provide inherently interpretable decision-making processes, and learn non-linear relationships effectively. Moreover, DTs often serve as the base learners in powerful ensemble methods like random forests and gradient boosting, which combine multiple DTs to create robust predictive models.

The concerns about feasible predictions become even more critical when a predictive model is integrated into an optimization framework. In these cases, the model’s predictions directly influence downstream decision-making. If the predictions violate known constraints, the optimization problem may become infeasible, making the entire decision-making process ineffective. Incorporating predictive models or their outputs into optimization is known as constraint learning \citep[\textit{e.g.},][]{FAJEMISIN20241} or predict-then-optimize \citep[\textit{e.g.},][]{wilder2019melding}, respectively. In both approaches, machine learning models must not only be accurate but also generate outputs that adhere to the constraints necessary for effective decision-making.

To illustrate the importance of adhering to such output constraints and making feasible predictions, we will provide two motivating examples that show how predictions lacking these qualities are not useful in practice.

\noindent\textbf{Hierarchical datasets. } Here, we briefly discuss two prominent applications where a predetermined hierarchy exists among output variables. Hierarchical time series (HTS) forecasting is a common application in inventory management, where predictions at different time scales must align with each other. In HTS, time series data is organized within a hierarchical aggregation structure, where values at lower levels (\textit{e.g.}, weekly demand) must aggregate to match those at higher levels (\textit{e.g.}, quarterly or yearly demand). For instance, when forecasting product demand, we need to ensure that the sum of predicted weekly demands equals the quarterly demand forecast, which in turn must align with the annual forecast. Standard decision trees cannot maintain these hierarchical constraints. When a DT predicts weekly demand independently of quarterly or annual forecasts, aggregated predictions often violate the hierarchical relationships inherent in the data, resulting in forecasts that fail to maintain the required hierarchical consistency.

In a similar vein, Hierarchical Multi-label Classification (HMC) presents a parallel challenge where rigid structural dependencies govern the prediction of class labels. In HMC, the target variables are organized into a taxonomy—typically a tree or a Directed Acyclic Graph—where a fundamental hierarchy constraint dictates that the assignment of a specific label necessitates the assignment of all its ancestor labels \citep{vens2008decision}. A prominent application is automated medical diagnosis, where diseases are classified according to standardized systems. For example, if a patient record is tagged with the specific diagnosis ``Viral Pneumonia,'' it must logically also satisfy the criteria for the parent category ``Pneumonia'' and the broader class ``Respiratory System Diseases.'' Standard classification algorithms often treat these diagnostic labels as independent targets, ignoring the medical taxonomy. This can lead to clinically incoherent results, such as a model predicting the presence of ``Viral Pneumonia'' while simultaneously predicting the absence of ``Pneumonia,'' thereby violating the logical entailment required by the diagnostic hierarchy.

\noindent\textbf{Retail logistics. } The second example is about retail logistics, where we manage a set of products--each with distinct attributes--in a central warehouse that will be delivered to retail stores. The goal is to predict the amounts of shipment for the products across these stores. However, due to budget or capacity restrictions, a store can only transport a product to a maximum of a fixed number out of the total stores. If we use the standard mean vector approach for prediction (averaging the target vectors within a leaf node), the resulting prediction may violate  both the integrality constraint and the capacity constraint by suggesting distribution to more than the allowed number of stores. Consequently, if this prediction is fed into a logistics optimization problem, it would result in an infeasible solution, rendering the entire decision-making process ineffective.

\begin{figure}
  \centering
  \includegraphics[scale=0.4]{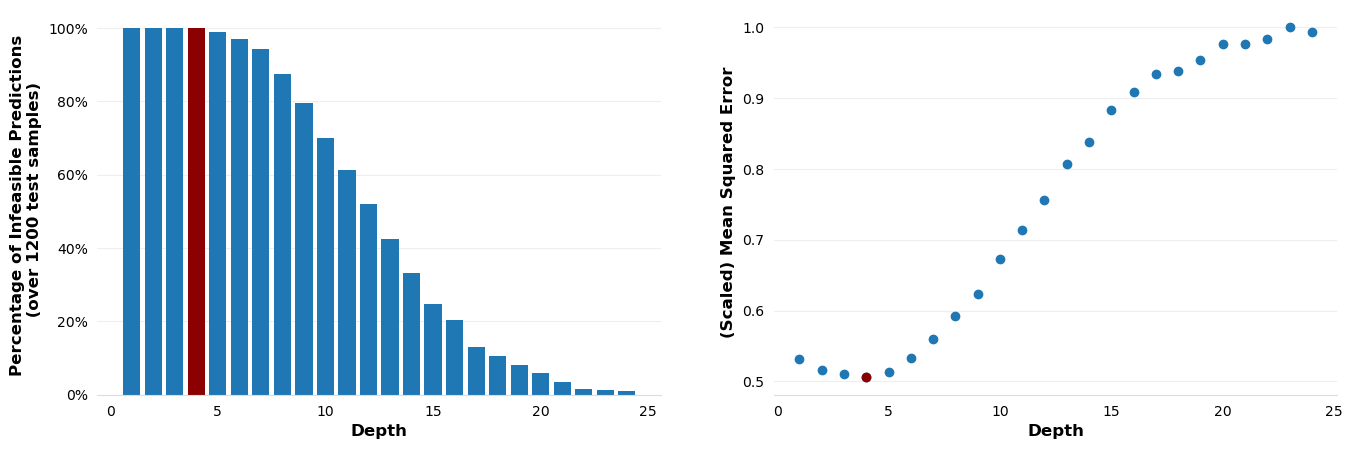}
  \caption{The trade-off between accuracy and feasibility. The red circle in the right plot marks the depth of the tree which returns the lowest mean squared error value for the test set. The corresponding red bar in the left plot shows the corresponding percentage of infeasible predictions. The sample size is 4000.}
  \label{fig:motivating_ex}
\end{figure}

To achieve feasible predictions, several ad-hoc alternatives can be suggested. First, for a given leaf node, it is possible to choose a representative instance as the prediction, provided that all training targets are already feasible. However, this approach fails when the training set is subject to noise. It is very common to encounter noise during data collection, meaning that even training instances might violate the underlying constraints. A practical example of this is found in weather forecasting using sensor data. It is not uncommon for sensors to be of low quality, resulting in unreliable data that introduces an additional layer of uncertainty \citep{chakraborty2020statistical}. Furthermore, inaccuracies in record-keeping are prevalent as well, often leading to inconsistent or even unrealistic data entries \citep[\textit{e.g.},][]{goldberg2008analysis,dehoratius2008inventory}. Second,  one might suggest training a deeper DT that would somehow guarantee feasibility by creating more homogeneous leaf nodes. However, this approach introduces a critical trade-off between accuracy and feasibility. As shown in Figure \ref{fig:motivating_ex}, deeper trees may increase the likelihood of satisfying output constraints, but at the cost of potential overfitting and reduced generalization performance. Furthermore, deeper trees sacrifice the interpretability that makes DTs valuable in the first place.

We intentionally focus on multi-target regression trees. In contrast, for multi-target classification problems, when there are only a few classes for each target and a small number of targets, it is possible to treat each unique combination of target values as a single label. This approach effectively reduces the problem to a standard single-target classification problem, albeit possibly with many labels. However, this workaround is not applicable to regression problems, where the output space is continuous, or to classification problems with a large number of targets or classes per target.

In this paper, we make the following contributions:
\begin{sibitemize}
  \item We introduce new optimization-based methods that ensure the feasibility of DT predictions by incorporating target constraints into the splitting procedure of common binary tree construction methods.

  \item We analyze the conditions for special feasible sets and loss functions under which the resulting constrained prediction problem boils down to an unconstrained one that can be solved efficiently.

  \item We discuss the conditions under which ensemble models -- such as Random Forest, AdaBoost, and others -- can incorporate our proposed approaches to guarantee the feasibility of their predictions.

  \item We present a computational study to discuss the efficacy of our methodology. We also demonstrate the practical application of our models in the context of HTS forecasting, particularly to predict weekly demand of products that aligns with quarterly forecasts and annual demand.

  \item We share a reproducible computational study using our publicly available implementation, enabling researchers and practitioners to apply and extend our methods to various domains.
\end{sibitemize}

\section{Related Literature.}
\label{sec:literature}
Traditional supervised learning methods primarily focus on predicting a single output, using data comprised of instances represented by features and a single output/response variable. In contrast, multi-output learning seeks to predict multiple output variables for a given set of input instances. The inherent dependencies between these output variables, along with the goal of capturing complex interactions among them, make multi-output learning more intricate compared to single-output approaches.  A prominent example of this paradigm is HMC. This is a specialized classification task in which instances are assigned to multiple classes organized within a specific hierarchy \citep[\textit{e.g.},][]{vens2008decision,silla2011survey,wehrmann2018hierarchical}.

As machine learning methods improve in accuracy, they may still produce predictions and outcomes that lack physical consistency or feasibility, often due to biases or data sparsity \citep{ZHANG2022105100}. Specifically, purely data-driven models may fail to align with constraints, natural facts, or standards derived from prior knowledge \citep{brundage2020toward}. Consequently, incorporating prior knowledge into machine learning models, especially in multi-output learning, has emerged as a prominent research avenue. This knowledge can take various forms including algebraic \citep[\textit{e.g.},][]{stewart2017label,heese2019good,ramamurthy2019leveraging} and differential equations \citep[\textit{e.g.},][]{yang2018physics,raissi2019physics}, simulation results \citep[\textit{e.g.},][]{lee2018spigan,daw2022physics}, or logic rules \citep[\textit{e.g.},][]{xu2018semantic,sachan2018learning}. Also, prior knowledge can vary greatly in nature and application, but here, we focus specifically on knowledge that pertains to the relationships involving only the output variables. The remainder of this section delves into the critical role of integrating prior knowledge as constraints into machine learning models, highlighting studies that examine why and when this approach is essential.

There exist various approaches on how to integrate prior knowledge into machine learning \citep{von2021informed}. Among these, embedding prior knowledge into learning algorithms is particularly important when ensuring consistency between predictions and prior knowledge is essential -- as is often required in data-driven decision-making problems. In this context, hybrid methods that integrate optimization techniques into machine learning models have gained significant attention in recent years due to their ability to leverage the strengths of both approaches. Hybrid approaches often integrate prior knowledge as constraints that are expected to be satisfied consistently. One of the earliest hybrid approaches of imposed constraints was implemented by \cite{lee1990neural}, who introduced constraints to deep learning architectures to approximate ordinary and partial differential equations. This approach laid the foundation for the general framework known as physics-informed neural networks, proposed by \cite{raissi2019physics}, where constraints and boundary conditions are incorporated \citep{rudin2022interpretable}. Similarly, \cite{christopher2024constrained} integrated constraints and physical principles to generative diffusion processes.  In a similar vein, there are studies aiming to regularize the curve or surface being learned to ensure it adheres to physical or economic theory regarding the \textit{input-output} relationship. \citep[\textit{e.g.},][]{AitSahalia2003, Miller1985, Curmei2020, Aubin2020, Papp2014}  Readers interested in knowledge-informed machine learning are referred to a review paper by \citet{von2021informed}.

\citet{faghihi2023gluecons} divided constraint based knowledge integration approaches into two groups: inference-time integration \citep[\textit{e.g.},][]{roth2017integer,deutsch2019general,lee2019gradient} and training-time integration \citep[\textit{e.g.},][]{nandwani2019primal,asai2020logic,yang2021safe}. Inference-time integration imposes constraints after the training process is completed, effectively acting as a post-processing step. In contrast, training-time integration incorporates constraints directly during model training. Both have been extensively investigated mostly with neural networks. \citet{li2019logic} emphasize that training-time integration does not guarantee that the outputs will satisfy the imposed constraints for all test instances. Constraints are often integrated into machine learning algorithms by extending the loss function with additional terms, which, while beneficial, does not ensure feasibility. However, \citet{faghihi2023gluecons} highlight that this approach often leads to higher prediction accuracy and reduced computational complexity compared to inference-time integration.


The integration of machine learning into optimization and decision-making represents a significant research avenue, where constrained prediction and guaranteed feasibility are critical concerns. Within this context, two main research streams have emerged: (i) directly predicting solutions of optimization problems, and (ii) the optimization with constraint learning (OCL) framework. The former typically relies on post-processing steps to repair infeasibilities and is often highly tailored to the problem at hand \citep[\textit{e.g.},][]{cappart2023combinatorial, joshi2019efficient, li2018combinatorial,khalil2016learning}. Specifically, in OCL setting, ensuring prediction feasibility is essential, as infeasible predictions can render the overall optimization problem infeasible.


DTs, which are originally single target machine learning approaches, have a structure that lends itself well to formulation as mixed integer programming (MIP) models, making them amenable to optimization-based approaches. Consequently, a significant body of literature has emerged to develop mathematical programming formulations of DTs \citep[\textit{e.g.},][]{aghaei2024strong,bertsimas2017optimal,dunn2018optimal,verwer2017learning}. There are also studies developing dynamic programming based methods to build optimal DTs \citep[\textit{e.g.},][]{zhang2023optimal,van2023necessary,demirovic2022murtree,lin2020generalized}. Nevertheless, we hereby restrict our attention solely to studies adopting mathematical programming formulations as they are more suitable to extensions for generalized constrained settings. Notably, DTs offer some flexibility in enforcing constraints, and many hybrid DT approaches incorporate \emph{instance-level} constraints to bind a subset of instances in generating meaningfully combined results. For instance, the clustering tree algorithm by \cite{struyf2007clustering} integrates domain knowledge to impose must-link and cannot-link constraints. DT methods have also been developed to address adversarial examples, which can be considered instance-level constraints \citep{chen2019robust, calzavara2019adversarial}, or attribute-level constraints to assign several features to the predictors \citep{nanfack2022constraint}. \cite{ben1995monotonicity} was one of the first to apply \emph{attribute-level} constraints by imposing monotonicity among the model's predictors, and \cite{cardoso2010classification} extended this in an ordinal classification setting. Nevertheless, we observe that the aforementioned constraints applied in DTs primarily focus either on design parameters, such as tree size or accuracy thresholds, or on relationships between input and output variables, which are often relatively simple in nature.

One of the earliest methods for multi-target regression with decision trees was introduced by \cite{mrt_cart}. He extended the univariate recursive partitioning approach from Classification and Regression Trees (CART) \citep{breiman1984cart} to handle multiple responses, calling this method multi-response trees. These are built similarly to CART, but the node impurity is redefined as the sum of squared errors across all response variables. \cite{struyf2006constraint} developed a system for building multi-objective regression trees, which include constraints on tree size and accuracy, allowing users to balance interpretability and accuracy by limiting tree size or setting a minimum accuracy threshold.

To the best of our knowledge, existing constrained machine learning algorithms have primarily focused on input-output relationships, such as monotonicity or convexity. The existing limited number of output-constrained methods, on the other hand, typically address only specialized or simple constraints, often failing to guarantee feasible predictions regarding inter-target dependencies for test instances. Addressing more general and complex output constraints while ensuring such guarantees remains an open challenge.

\section{Background.} 
\label{sec:def}

Suppose that we have a training dataset $\CD = \{(\vx_i, \vy_i): i \in \CI\}$ where $\CI = \{1, \dots, n\}$. Each sample consists of an input vector $\vx_i \in \RR^p$ and a target (output) vector $\vy_i \in \RR^K$. For convenience, let $x_{ik}$ and $y_{ik}$ denote the $k$th element of the associated vector. Here, we remark that not all the target vectors in the training dataset necessarily satisfy feasibility constraints due to noise present in the data. That is, if we denote the \textit{feasible set} of target values by $\YY \subseteq \RR^K$, then for some $i \in \CI$, we have $\vy_i \notin \YY$.
Our goal is to produce a prediction $\hat{\vy}_0$ that is both accurate and feasible for any given test sample $\vx_0$, \textit{i.e.}, $\hat{\vy}_0 \in \YY$. In the following discussion, we first discuss the challenges of ensuring feasibility in test instances for prediction models and argue that decision trees are uniquely positioned to address constrained prediction. Additionally, we provide some background information on traditional unconstrained decision trees and explore the practical important implications of output constraints on solution methods.

\noindent\textbf{Constrained Prediction. } When learning a model, we aim to find a function that can make accurate predictions with minimal error when compared to the target values (labels). In our constrained setting, however, not only do those models need to accurately capture the underlying structure of the data, but they also need to ensure that their predictions remain feasible. To achieve this, we can formulate the following constrained optimization problem
\[
  \min_{f\in \mathcal{F}} \left\{\sum_{i \in \CI} \ell(f(\vx_i), \vy_i) : f(\vx_i) \in \YY, i \in \CI\right\},
\]
where $\ell:\RR^K \times \RR^K \mapsto \RR$ is the loss function, $f:\RR^p \mapsto \RR^K$ is the predictive model, and $\mathcal{F}$ is the family of candidate models, also known as hypothesis class.

Solving this problem during training yields an optimal function $f^*$, but there is no guarantee that the predictions for test samples will remain feasible. That is, we cannot make certain that $f^*(\vx_0) \in \YY$ holds for every given test sample $\vx_0$. This limitation exists as the feasibility constraint is enforced only during training, and there is no explicit mechanism, in general, to guarantee that the resulting model will always produce feasible predictions for unseen test data.

However, decision trees offer a property that inherently preserves feasibility for test instances if feasibility is enforced during training. A decision tree partitions the input space into disjoint regions corresponding to the leaves of the tree. Each sample in the training dataset falls into one leaf, and the decision tree assigns the same prediction for all training samples in the leaf. More importantly, this prediction also applies to any given test sample. Therefore, if the predictions of the leaf nodes are all feasible, then we obtain feasible predictions also for the unseen data by construction. This remarkable property makes decision trees uniquely suited for output-constrained prediction.



\noindent\textbf{Decision Trees. } A regression/classification tree is grown by splitting the dataset into non-overlapping subsets each represented by a node in the tree. The most common approach for the tree construction employs binary splits; that is, each branching populates at most two child nodes, left ($\leftharpoonup$) and right ($\rightharpoonup$), derived from a parent node. Let $\CN \subseteq \CD$ denote the subset associated with the parent node, and $\CI_{\CN} \subseteq \CI$ represent the indices of the samples within that node. To generate a pair of child nodes, a feature $j$ and a corresponding split value $v$ are chosen to obtain the left and right index sets $\CI^\leftharpoonup_{\CN}(j, v) = \{i \in \CI_{\CN}: x_{ij} < v\}$ and $\CI^\rightharpoonup_{\CN}(j, v) = \{i \in \CI_{\CN}: x_{ij} \geq v\}$, respectively. Throughout the paper, we follow such a univariate and binary split convention.

A widely adopted approach for constructing such trees is recursive partitioning, where relatively simple and myopic prediction problems are solved at each branch node \citep{breiman1984cart}. The core idea is to evaluate a given split on feature $j$ at value $v$ by
\begin{equation}
  \label{eqn:gain}
  \mathcal{F}(j,v) = \alpha^\leftharpoonup_{\CN}(j, v) \sum_{i \in  \CI^\leftharpoonup_{\CN}(j, v)} \ell(\vy^\leftharpoonup_{\CN}, \vy_i)  + \alpha^\rightharpoonup_{\CN}(j, v) \sum_{i \in  \CI^\rightharpoonup_{\CN}(j, v)} \ell(\vy^\rightharpoonup_{\CN}, \vy_i),
\end{equation}
where $\vy^\leftharpoonup_{\CN}$ and $\vy^\rightharpoonup_{\CN}$ are the predictions of the left and right child nodes, respectively. Also, $\alpha^\leftharpoonup_{\CN}(j, v) = \tfrac{|\CI^\leftharpoonup_{\CN}(j, v)|}{|\CI_{\CN}|}$ and $\alpha^\rightharpoonup_{\CN}(j, v) = 1 - \alpha^\leftharpoonup_{\CN}(j, v)$ are the scaling coefficients representing the proportional weights of the corresponding subsets. As such, it first requires computing the optimal predictions $\vy^\leftharpoonup_{\CN}$ and $\vy^\rightharpoonup_{\CN}$ for a given split on feature $j$ at value $v$.

Computing the optimal predictions, which we refer to as the \textit{prediction problem}, is a fundamental component of decision tree construction. The output constraints have a direct impact on the prediction problem. For a given subset of training data $\CS$, the prediction problem can be expressed as follows:
\begin{equation} \label{eq:ConsPred}
  \vy_{\CS} = \arg \min_{\hat{\vy}\in \YY}\sum_{i \in \CS} \ell(\hat{\vy}, \vy_i).
\end{equation}
Depending on how the feasible set $\YY$ is defined, the resulting prediction problem may become difficult to solve optimally. For instance, a non-convex feasible set $\YY$ could pose substantial difficulties in solving the constrained prediction problem. This, in turn, complicates the construction of output-constrained decision trees. Next, we introduce several approaches to construct output-constrained regression trees (\ocrt s).

\section{Methodology.}
\label{sec:method}

The output-constrained regression tree differs significantly from its unconstrained counterpart as the resulting prediction problem can be considerably more challenging. It is well established that computing the optimal unconstrained regression/classification tree is an $\mathcal{NP}$-hard problem \citep{laurent1976constructing}. Incorporating output constraints only amplifies this computational challenge even further. As discussed earlier, mixed integer programming (MIP)  formulations for computing the optimal regression trees have been developed \citep[\textit{e.g.},][]{bertsimas2017optimal,dunn2018optimal}. Extending those models to accommodate vector-valued constrained outputs would also be straightforward. However, the MIP models, particularly for the optimal regression tree, are known to have weak linear programming relaxations, leading to poor computational efficiency \citep{aghaei2024strong}. Therefore, solving the MIP formulation to determine the optimal output-constrained regression tree is computationally prohibitive even for small datasets.

Given these limitations, we propose alternative approaches inspired by the CART framework, where splitting and prediction decisions are made recursively at each branch node. Nevertheless, they differ on how to address the splitting problem. First, we introduce an iterative heuristic that leverages an MIP model at each node to determine the locally optimal split and predictions in concert. Additionally, we develop an alternative heuristic that, akin to CART, employs an exhaustive search for locally optimal splits and addresses the constrained prediction problem optimally.

The proposed methods aim to solve the following optimization problem, differing primarily in how the split decisions—specifically, the feature and threshold that define the partition—are determined. For a given node with sample indices $\CI_{\CN}$, this problem aims to find the split parameters $(j, v)$ and child node predictions $\vy^\leftharpoonup_{\CN}, \vy^\rightharpoonup_{\CN}$ that minimize the total loss while satisfying the output constraints:
\begin{align}
  \min_{j, v, \vy^\leftharpoonup_{\CN}, \vy^\rightharpoonup_{\CN}} \left\{ \sum_{i \in \CI_N^\leftharpoonup(j,v)} \ell(\vy^\leftharpoonup_{\CN}, \vy_i) + \sum_{i \in \CI_{\CN}^\rightharpoonup(j,v)} \ell(\vy^\rightharpoonup_{\CN}, \vy_i): \vy^\leftharpoonup_{\CN}, \vy^\rightharpoonup_{\CN} \in \YY\right\}.
\end{align}

In the remainder of this section, we first introduce the proposed solution approaches for \ocrt. Then, we present the implications of the convex feasible set, which is an important, as well as common, special case. Finally, we discuss how to use \ocrt approaches in ensemble learning for constrained prediction purposes.

\subsection{MIP-based Training}
\label{subsubsec:IORT}

Here, the idea is to solve an MIP model at each node to determine locally optimal prediction and split decisions. Building on the well-established formulation from \citet{dunn2018optimal} for optimal regression trees, we slightly extend it to construct an output-constrained yet single-depth decision tree. This MIP model aims to minimize the total loss of two resulting child nodes. Like CART, this method recursively partitions the dataset by making sequential decisions at each node. However, unlike CART, which relies on enumerative approaches to determine splits, this method solves an MIP model at each node to optimize both the split and prediction decisions. As such, it eliminates the need for an exhaustive search to determine the optimal split.

The complete MIP formulation is given as
\[
  \begin{array}{ll}
    \minimize & \sum_{i\in \CI_{\CN}} \ell(\hat{\vy}_{i}, \vy_{i})                                                 \\
    \subto    & -M(1-z_{i}) \leq \hat{\vy}_{i} - \vy^\rightharpoonup_{\CN} \leq M(1-z_{i}), \quad  i\in \CI_{\CN}, \\
              & -Mz_{i} \leq \hat{\vy}_{i} - \vy^\leftharpoonup_{\CN} \leq Mz_{i} , \quad  i\in \CI_{\CN},         \\
              & \va^\top\vx_{i} \geq b - M(1-z_{i}), \quad  i\in \CI_{\CN},                                        \\
              & \va^\top\vx_{i} + \epsilon \leq b + Mz_{i}, \quad  i\in \CI_{\CN},                                 \\
              & N_{\min}\leq \sum_{i\in \CI_{\CN}}z_{i} \leq n-N_{\min},                                           \\
              & \ve^\top\va = 1, \quad  \va\in\{0,1\}^p, \quad b \in \mathbb{R},                                   \\
              & z_{i} \in \{0,1\}, \quad \hat{\vy}_i \in \mathbb{R}, \quad i\in \CI_{\CN},                         \\
              & \vy^\leftharpoonup_{\CN}, \vy^\rightharpoonup_{\CN} \in \YY,
  \end{array}
\]
where $\hat{\vy}_i$ denotes the prediction vector for instance $i$, $z_{i}$ is a binary variable indicating if instance $i$ is assigned to the right node, $\va$ represents binary variables that indicate if a split is applied on a particular feature, $b$ denotes the split threshold applied at the parent node, $M$ represents a sufficiently large constant, $\epsilon$ denotes a constant to convert strict inequalities into weak inequalities, and $N_{\min}$ refers to the minimum number of instances required in each leaf node.

Specifically, the proposed approach operates as follows. Starting from the root node, this approach applies the MIP model to compute the optimal constrained single-depth regression tree at each node. For any given parent node, the model determines the optimal split location and the corresponding predictions for the two resulting child nodes. The total loss obtained from this split is then compared to the loss associated with the parent node. If the split leads to a reduction in loss, the split is accepted, and the process continues recursively for the child nodes. If not --\textit{i.e.}, if the split increases the loss-- the parent node is not a branch node and is instead designated as a leaf node. This process continues iteratively until all parent nodes have been processed. Consequently, the MIP model must be solved at each branch node throughout the tree-growing process. This approach can be far more time efficient than solving the MIP formulation for the entire tree at once. Throughout the paper, this method is referred to as \mocrt.


\subsection{Enumerative Training}
\label{subsubsec:EORT}

The approach introduced here follows the CART methodology in constructing the regression tree. Specifically, it recursively partitions the dataset and grows the tree by evaluating \eqref{eqn:gain} at all possible splits corresponding to feature-threshold pairs. For each candidate split, the resulting constrained prediction problem~\eqref{eq:ConsPred} is solved using a dedicated optimization method. We shall refer to this method as \eocrt.

Clearly, solving an optimization problem for each feature-threshold pair may pose computational challenges and increase the training time when the problem dimension is large. Fortunately, in many applications, the dimension of the prediction problem ($K$) is relatively small, and hence, modern optimization solvers can obtain solutions very fast. The following proposition establishes when the trees obtained by \mocrt and \eocrt are identical. The proof of this proposition follows from the definitions of the two approaches and the loss functions.

\begin{proposition}
  \label{lem:SDRT}
  When \eqref{eqn:gain} serves as the objective function for \eocrt and the loss function is the average error over the samples (\textit{e.g.}, MSE or MAD), \mocrt\ and \eocrt\ yield the same regression tree.
\end{proposition}

Despite this result, the computational performances of \mocrt and \eocrt may differ significantly. As \mocrt relies on a relatively large MIP formulation, it is expected to scale less efficiently than \eocrt for large datasets. Consequently, \mocrt may be a more computationally viable option for small- to medium-sized datasets, whereas \eocrt is likely better suited for larger-scale problems.

A practical special case of \eocrt can be introduced by relaxing the requirement that the constrained prediction problem be solved at every candidate split. In this variant, the tree is first constructed by solving the unconstrained prediction problem--following the standard CART procedure--and the constrained prediction problem is subsequently applied only at the leaf nodes of the resulting tree. This approach, referred to as \epocrt, can also be viewed as a post-processing method because it does not account for output constraints during tree construction but repairs prediction infeasibility at the leaf level. It is important to note that \epocrt, similar to \eocrt and \mocrt, guarantees the feasibility of the final predictions. As such, \epocrt may offer a computationally viable alternative to \eocrt and \mocrt, albeit potentially at the expense of predictive performance.

\subsection{Convex Feasible Set}
\label{sec:convex}
In practice, convex feasible sets--often defined by simple bounds or, more generally, linear constraints on the target vectors--are very common. A prominent example is hierarchical forecasting where predictions across different levels--whether based on time, geography, or other structured dimensions--must be consistent with each other. Specifically, predictions at lower levels must aggregate to match those at higher levels, ensuring coherence within the hierarchy  (see also Section \ref{sec:intro}).
Additionally, many real-world scenarios require predictions to fall within specific bounds, often due to capacity constraints or resource limitations. As discussed above, these constraints naturally lead to convex feasible sets.

The structure of the feasible set significantly influences the complexity of the constrained prediction problem~\eqref{eq:ConsPred} to be solved while constructing \ocrt. In this regard, when the feasible set is convex, the problem~\eqref{eq:ConsPred} becomes more tractable, particularly if the objective function is also convex--as it is often the case with loss functions used in prediction tasks. In such cases, convexity eliminates the risk of suboptimal local minima enabling methods such as gradient descent, interior-point, and simplex algorithm to efficiently compute the global optimum.

Here, we take this analysis a step further by considering another important special case related to the characteristics of the dataset: a noise-free setting where all target vectors in the training set are feasible, \textit{i.e.}, $\vy_i\in \YY$ for all $i\in \CI$.  In this scenario, the constrained prediction problem~\eqref{eq:ConsPred} can be addressed using existing tree-growing heuristics. Specifically, since every training target is already feasible, one can select a representative point--such as the medoid or centroid--among the relevant subset of the training set, thereby simplifying the solution process.

Consider the case where the feasible set is convex and the dataset is noise-free. Recall that for loss functions such as mean squared error and mean Poisson error, the mean of the target vectors within a given training subset provides the optimal solution for the unconstrained prediction problem. Similarly, when employing mean absolute loss, the median output vector is the optimal solution.

The following theorem establishes that, under a convex feasible set and a noise-free dataset, the mean vector remains optimal for mean squared loss and mean Poisson loss. The proof follows from the observation that the mean operation corresponds to taking a convex combination of the feasible target vectors within a node. A formal proof of this result is provided in Section \ref{prf:theorem} of the appendix.

\begin{theorem} \label{thm1}
  Let $\YY \subseteq \RR^k$ be convex and $\vy_i\in \YY$ for all $i\in \CI$. If the loss function is the mean squared error or the mean Poisson deviance, then the mean of the target vectors in a node minimizes the loss.
\end{theorem}

This theorem suggests that for a restricted class of feasible sets combined with noise-free output, standard decision tree implementations can be used effectively with certain loss functions.  However, this result does not extend to cases where the feasible set is non-convex or the target vectors are subject to noise. Therefore, we need our proposed approaches $\mocrt$ or $\eocrt$.

\subsection{Ensemble Training} \label{sec:Ensemble}
Ensemble learning methods, such as random forests and tree-based boosting techniques, generate predictions by taking weighted combinations of base (weak) learners, which serve as the fundamental building blocks of the ensemble \citep{breiman2001random}. In general, even if each base learner produces a feasible prediction, their weighted combination does not necessarily preserve feasibility.

Nevertheless, a notable special case here arises when the feasible set is convex, and predictions are obtained through convex combinations of the base learners' outputs. In this setting, \mocrt or \eocrt can be incorporated into ensemble training to ensure feasibility even if the output is noisy. The following proposition formally establishes this observation and highlights one of the key findings of this section.
\begin{proposition} \label{rem:Ensemble}
  When the feasible set $\YY \subseteq \RR^k$ is convex, ensemble learning methods that employ \mocrt or \eocrt as base learners, and generate predictions through their convex combinations
  also guarantee feasibility for test instances.
\end{proposition}

This proposition can be easily proved based on two observations: (i) \mocrt and \eocrt always provide feasible predictions, ensuring that each base learner's output lies within the feasible set, and (ii) the weighted average of feasible predictions corresponds to a convex combination of those predictions, which, due to the convexity of the feasible set, guarantees that the resulting prediction remains within the set.

This is an important finding as it paves the way for more sophisticated constrained prediction methods under convex feasible sets. Ensemble methods are widely recognized for enhancing the performance of base (weak) learners, making them valuable tools for improving prediction accuracy. The ability to integrate ensemble methods into constrained learning frameworks leads to more precise and efficient constrained predictions. Ensemble methods that can be employed in this context include Random Forest, AdaBoost R2, LogitBoost, BrownBoost, TotalBoost, and SGD Boosting (with normalization).

\section{Computational Study.}
\label{sec:comp}

This section evaluates the performance of the proposed methods, \mocrt, \eocrt, and \epocrt, for training output-constrained regression trees (\ocrt) and their corresponding random forest implementations, \mrf, \erf, and \prf. As a benchmark, we consider the standard regression tree trained via CART. In this study, we use the \texttt{scikit-learn} implementation, denoted as \skl, solely for generating the standard regression trees \citep{scikit-learn}.

The computational study is organized in two parts, each serving a distinct purpose. First, we assess the predictive accuracy and computational efficiency of the proposed approaches in a controlled setting using randomly generated synthetic datasets. Second, we demonstrate the practical applicability of constrained prediction through a real-life case study inspired by specific operational challenges commonly encountered in hierarchical forecasting across sectors such as retail demand planning, energy load forecasting, transportation management, and financial budgeting, where predictions must remain coherent across multiple aggregation levels. All computational experiments are reproducible by using the dedicated GitHub repository\footnote{\url{https://github.com/sibirbil/OCDT}}. Numerical experiments are conducted on an Intel i7-10510u 1.80GHz processor and use Gurobi 11.0.1 as the MIP solver. For \mocrt and \mrf methods, we impose a time limit of 120 seconds for each MIP model solved at each decision tree split. Also, unless stated otherwise, we use MSE for split evaluation and leaf prediction in all experiments. Moreover, we conduct further experiments to explore the computational limits of our methods and analyze the impact of problem-specific characteristics. The results of those further experiments are presented in Appendix~\ref{app:experiments}.  Additional implementation details are provided in the following sub-sections.


\subsection{Performance Tests on Synthetic Datasets}

In this part of the computational study, we aim to evaluate the performance of \mocrt, \eocrt, \epocrt, \erf, and \prf against \skl. Recall that random forest implementations  \erf, and \prf can guarantee feasible predictions only when the feasibility set is convex. Hence, we generate synthetic datasets under convex feasibility assumption. Consequently, all methods except \skl --namely \mocrt, \eocrt, \epocrt, \erf, and \prf--ensure feasibility of predictions. It is also worth noting that although we include \mrf --\textit{i.e.}, the \rf implementation of \mocrt-- in our numerical experiments, we do not report its results. For relatively small instances where \mocrt can be solved within reasonable time limits, \mrf yields results identical to those of \erf. However, for larger datasets, solving the underlying MIP problems becomes computationally intractable. Therefore, we omit \mrf from the reported results.

For all regression tree methods, we experiment with two maximum tree depths of five and seven, while ensuring a minimum of ten data points to split a node and at least five data points in each leaf. All random forest implementations use ensembles of 20 regression trees. To measure the performance of the proposed methods, we generate synthetic datasets with varying sizes $n\in\{500,1000,2000\}$ and numbers of targets $K\in\{5,9\}$. For each combination of dataset size and number of targets, we generate three random datasets. We use a full factorial design and this leads us to 18 datasets in total. Each dataset contains a fixed number of input features, \textit{i.e.}, $p=6$. We conduct a five-fold cross-validation on each dataset using two different tree-depth parameter values for each method listed above.

The synthetic data generation process is given as follows: Each feature $X$ is drawn uniformly from a discrete set of 11 values evenly spaced on the interval $[0,1]$. For each target variable $k$, a functional relationship is constructed using a randomly selected combination of six function types: sine, cosine, tangent, linear, and polynomial. Specifically, for each target $k$, we generate a function of the form,
\[
  y_{ik}=\sum_{j=1}^6w_{jk}f_{jk}(x_{ij};\theta_{jk})
\]
where, for each target $k$, $f_{jk}$  is a function randomly selected from the specified set and applied to feature $x_{ij}$ with randomly sampled parameters $\theta_{jk}$. The coefficient $w_{jk}$ denotes a randomly chosen strength multiplier for the corresponding function $f_{jk}$.

At this point, one can observe that the predetermined constraints do not have any impact on the generated target values. To ensure compliance with these constraints to a certain extent, we apply a simple transformation to each target vector $\vy_i$. Specifically, we project each resulting target vector onto a polyhedron. Accordingly, we design the constraints to conform to this affine form. Notably, we leave out the non-negativity constraints from the projection step in order to deliberately introduce a layer of noise. As a result, while the transformed target vectors satisfy all constraints included in the projection, they may violate the non-negativity constraints that the original target values were expected to satisfy. We finally add a random noise term to the transformed target vectors to introduce an additional layer of uncertainty. The constraints employed in synthetic datasets are provided explicitly in Appendix~\ref{app:synthetic_optimization}.

Tables~\ref{tab:syn_avg}–\ref{tab:syn_avg_depth} present the results obtained from synthetic datasets. For convenience, \ocrt and \rf methods are presented in different columns. For each method, the tables report the average percentage mean squared error (MSE) gap, denoted by $\Delta$, relative to \skl. Specifically, a positive (negative) value of $\Delta$ indicates that the method’s MSE is higher (lower) than that of \skl, thereby quantifying the relative performance of each approach. It is important to highlight that \skl provides infeasible predictions in all our tests.

\begin{table}[htbp]
  \centering
  \small
  \caption{Average MSE gaps on synthetic datasets}
  \begin{tabular}{lr|lr}
    \toprule
    \ocrt Methods & \multicolumn{1}{c|}{$\Delta$} & \rf Methods & \multicolumn{1}{c}{$\Delta$} \\
    \midrule
    \epocrt       & 62.02\%                       & \prf        & 57.75\%                      \\
    \eocrt        & 59.44\%                       & \erf        & 55.10\%                      \\
    \mocrt        & 62.31\%                       &             &                              \\
    \bottomrule
  \end{tabular}%
  \label{tab:syn_avg}%
\end{table}%

Table~\ref{tab:syn_avg} reports the average percentage MSE gaps for each method across all synthetic datasets. As expected, \rf methods outperform their \ocrt counterparts. Among the \ocrt methods, \eocrt achieves the lowest average MSE gap. Although \eocrt and \mocrt are designed to produce identical trees, differences arise here due to the computation time limits imposed on the MIP models in \mocrt, resulting in suboptimal splits and thus lower performance. Interestingly, \epocrt performs nearly as well as \mocrt, despite not incorporating output constraints during tree construction. However, it still falls behind \eocrt by approximately 3\%. Unsurprisingly, \erf yields better performance than \prf.

\begin{table}[htbp]
  \centering
  \small
  \caption{Average MSE gaps on synthetic datasets for different numbers of instances}
  \begin{tabular}{clr|lr}
    \toprule
    \multicolumn{1}{l}{$n$}  & \ocrt Methods & \multicolumn{1}{c|}{$\Delta$} & \rf Methods & \multicolumn{1}{c}{$\Delta$} \\
    \midrule
    \multirow{3}[2]{*}{500}  & \epocrt       & 52.75\%                       & \prf        & 46.14\%                      \\
                             & \eocrt        & 49.74\%                       & \erf        & 43.51\%                      \\
                             & \mocrt        & 50.16\%                       &             &                              \\
    \midrule
    \multirow{3}[2]{*}{1000} & \epocrt       & 61.51\%                       & \prf        & 57.91\%                      \\
                             & \eocrt        & 59.85\%                       & \erf        & 55.59\%                      \\
                             & \mocrt        & 60.64\%                       &             &                              \\
    \midrule
    \multirow{3}[2]{*}{2000} & \epocrt       & 71.80\%                       & \prf        & 69.18\%                      \\
                             & \eocrt        & 68.73\%                       & \erf        & 66.21\%                      \\
                             & \mocrt        & 76.14\%                       &             &                              \\
    \bottomrule
  \end{tabular}%
  \label{tab:syn_avg_size}%
\end{table}%

Table~\ref{tab:syn_avg_size} presents the average percentage MSE gaps for each method across all synthetic datasets characterized with varying data sizes. Among the \ocrt methods, \eocrt consistently emerges as the best approach. \mocrt performs comparably to \eocrt on relatively small datasets; however, as the dataset size increases, \epocrt begins to outperform \mocrt, since solving the underlying MIP formulation of \mocrt becomes increasingly difficult. A similar trend can be observed for the \rf methods: \erf consistently outperforms \prf in all instances.

\begin{table}[htbp]
  \centering
  \small
  \caption{Average MSE gaps on synthetic datasets for different numbers of targets}
  \begin{tabular}{clr|lr}
    \toprule
    \multicolumn{1}{l}{$K$} & \ocrt Methods & \multicolumn{1}{c|}{$\Delta$} & \rf Methods & \multicolumn{1}{c}{$\Delta$} \\
    \midrule
    \multirow{3}[2]{*}{5}   & \epocrt       & 56.73\%                       & \prf        & 56.80\%                      \\
                            & \eocrt        & 54.88\%                       & \erf        & 53.87\%                      \\
                            & \mocrt        & 56.50\%                       &             &                              \\
    \midrule
    \multirow{3}[2]{*}{9}   & \epocrt       & 67.31\%                       & \prf        & 58.69\%                      \\
                            & \eocrt        & 64.00\%                       & \erf        & 56.34\%                      \\
                            & \mocrt        & 68.12\%                       &             &                              \\
    \bottomrule
  \end{tabular}%
  \label{tab:syn_avg_target}%
\end{table}%

Table~\ref{tab:syn_avg_target} reports the average percentage MSE gaps for each method across all synthetic datasets with varying numbers of target variables. As before, \eocrt consistently emerges as the best-performing \ocrt method. Interestingly, the performance gap between \eocrt and the other methods tends to escalate as the number of targets increases. This is likely because additional targets introduce greater combinatorial or optimization-wise complexity, thereby amplifying the benefits of integrating output constraints during model construction. In contrast, \mocrt is outperformed by both alternatives, particularly when the number of targets is large. Among the \rf methods, \erf outperforms \prf, although there is no clear trend with respect to the number of targets.

\begin{table}[htbp]
  \centering
  \small
  \caption{Average MSE gaps on synthetic datasets for different tree depths}
  \begin{tabular}{clr|lr}
    \toprule
    \multicolumn{1}{l}{Depth} & \ocrt Methods & \multicolumn{1}{c|}{$\Delta$} & \rf Methods & \multicolumn{1}{c}{$\Delta$} \\
    \midrule
    \multirow{3}[2]{*}{5}     & \epocrt       & 57.00\%                       & \prf        & 53.04\%                      \\
                              & \eocrt        & 54.53\%                       & \erf        & 50.35\%                      \\
                              & \mocrt        & 57.74\%                       &             &                              \\
    \midrule
    \multirow{3}[2]{*}{7}     & \epocrt       & 67.04\%                       & \prf        & 62.45\%                      \\
                              & \eocrt        & 64.36\%                       & \erf        & 59.86\%                      \\
                              & \mocrt        & 66.88\%                       &             &                              \\
    \bottomrule
  \end{tabular}%
  \label{tab:syn_avg_depth}%
\end{table}%

Table~\ref{tab:syn_avg_depth} presents the average percentage MSE gaps for each method across all synthetic datasets for different tree sizes. Here, we can observe that tree size does not depict an observable impact on prediction performances of each method. However, it is observable that the overall performance of all constrained informed models yields relatively greater MSE values as compared to \skl. This is possibly due to the increasing impact of constraints on each presumably smaller subsets of data falling within each leaf node.

\begin{figure}[htbp]
  \centering
  \begin{subcaptionbox}{Average training time per method\label{fig:fig1}}[0.48\textwidth]
    {\includegraphics[width=\linewidth]{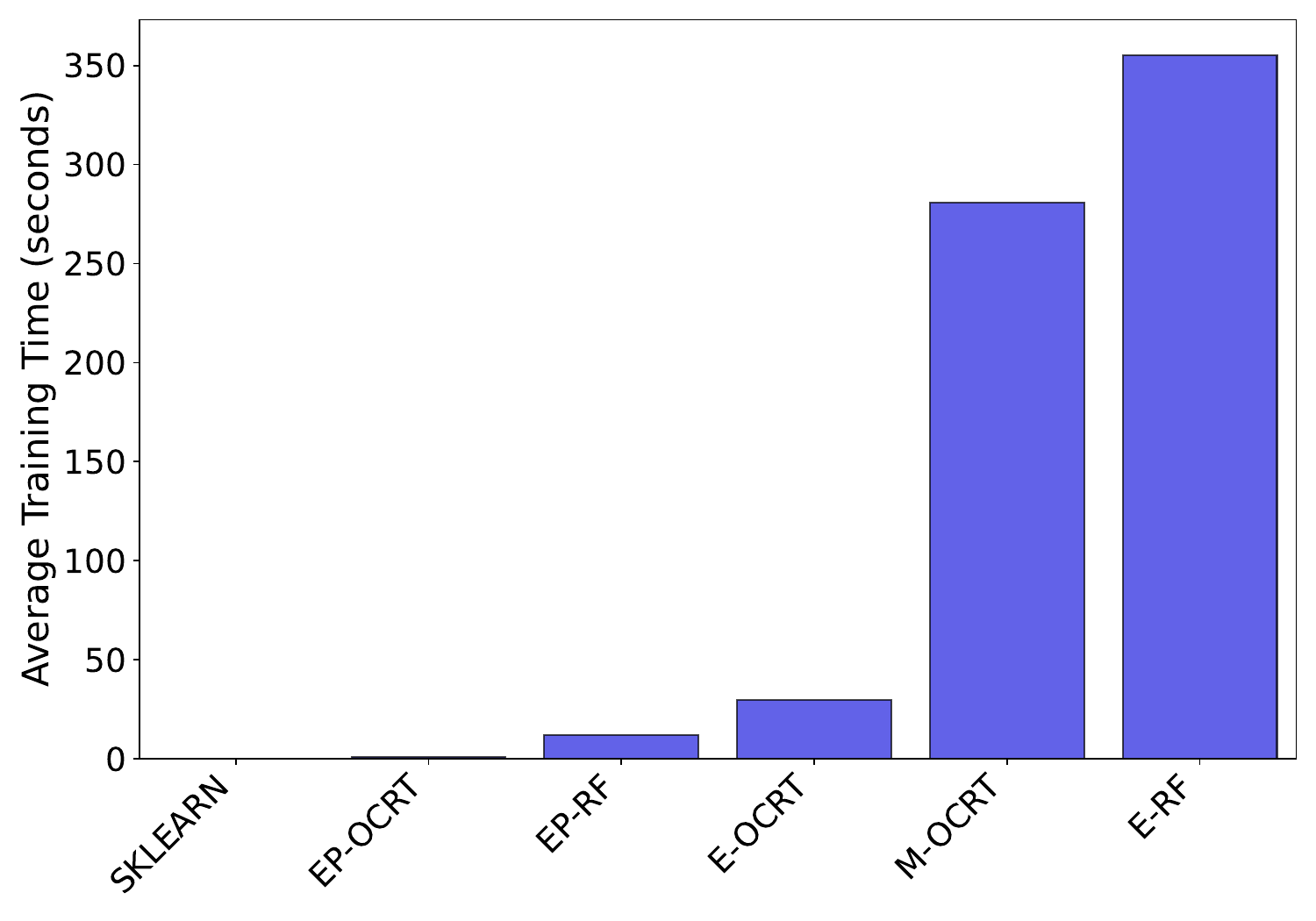}}
  \end{subcaptionbox}
  \hspace{0.009\textwidth}
  \begin{subcaptionbox}{Average training time (log scale) per method by size \label{fig:fig2}}[0.48\textwidth]
    {\includegraphics[width=\linewidth]{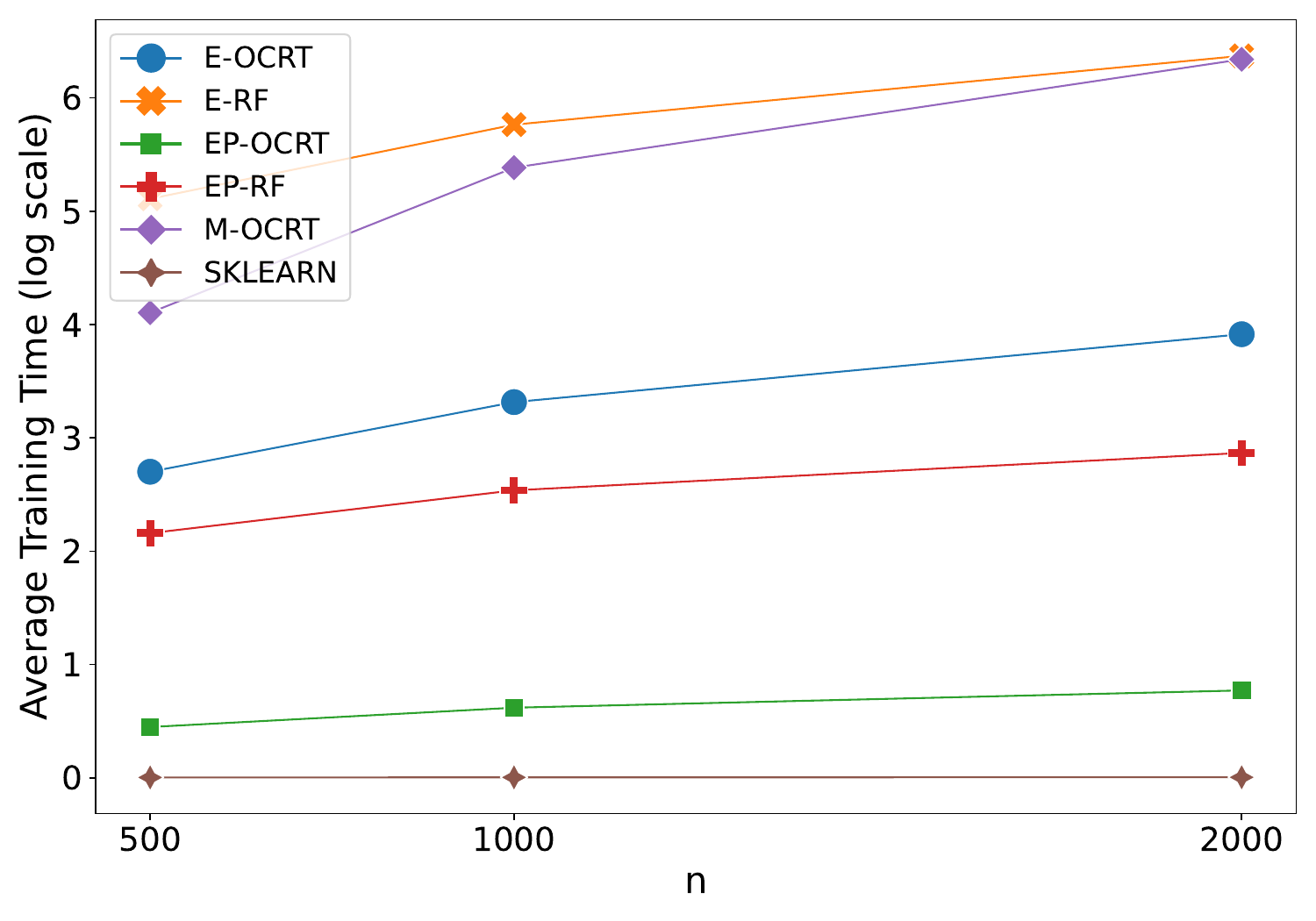}}
  \end{subcaptionbox}
  \caption{Average training times}
  \label{fig:ttime}
\end{figure}

Figure~\ref{fig:ttime} illustrates the trends in average computation times for each method. Specifically, the left panel (Figure~\ref{fig:fig1}) presents the overall average computation time per method across all datasets, while the right panel (Figure~\ref{fig:fig2}) shows how computation time scales with increasing data size. As seen in Figure~\ref{fig:fig1}, the \skl and \epocrt methods are the most time-efficient, as expected. The \prf and \eocrt methods also demonstrate relatively good computational performance. Among the \ocrt methods, \mocrt stands out as the most time-intensive by far. Interestingly, the computation times of \mocrt and \erf can be observed as somewhat comparable. Figure~\ref{fig:fig2} clearly shows that computation times of all methods increase with dataset size. This trend is especially pronounced for \mocrt and \erf and is also noticeable for \eocrt and \prf.

\subsection{Hierarchical Time Series}
A hierarchical time series (HTS) is a collection of time series data organized within a hierarchical aggregation structure, where values at lower levels aggregate to match those at higher levels. HTS models are widely used in industry to ensure consistency across interconnected forecasts.

A common application is inventory management, where weekly demand predictions must align with higher-level forecasts, such as quarterly or yearly demand. This is particularly relevant for products with sparse historical demand, where most observations are zero. In such cases, directly applying a regression model to predict weekly demand often results in naive predictions of zero due to the lack of meaningful patterns in the data.

To overcome this limitation, a two-step approach is often employed:
\begin{enumerate}
  \item Quarterly Demand Prediction: First, determine the demand for a given quarter.
  \item Conditional Weekly Demand Prediction: If quarterly demand is predicted to exist, a separate predictive model, trained only on instances where quarterly demand is greater than zero, is used to estimate weekly demand.
\end{enumerate}

HTS models are particularly valuable when different parts of a business rely on interconnected forecasts that must be consistent with each other. For example, the forecasted weekly demands may need to respect upper bounds from business constraints or higher-level forecasts. Similarly, the forecasted total demand for a region may need to equal the sum of forecasts for individual stores in that region.

In this section, we demonstrate how an \ocrt model can be used to predict weekly demand of stock keeping units (SKUs) that aligns with quarterly forecasts and adheres to existing operational constraints. Unlike traditional methods that involve post-processing predictions to enforce constraints, the \ocrt approach integrates these constraints directly into the modeling process. This ensures forecasts are coherent from the outset, avoiding the inefficiencies and inaccuracies that can arise from post-hoc adjustments.  As such, the  \ocrt implementation in this case study refers to obtaining constraint-aware predictions in intermediate steps of the two-step sequential prediction approach described above.

Here, we consider two sets of constraints: weekly demand aggregation constraints and sparsity constraints. The aggregation constraints ensure consistency with quarterly demand predictions. Furthermore, the sparsity constraints reflect the lumpy and intermittent nature of the demand, often common in spare parts.  These sparsity constraints render the feasible set inherently non-convex. We use five randomly generated datasets, each containing $n = 500$ instances, $p = 6$ input features, and $K = 13$ target variables. Furthermore, to assess the impact of noise on the predictive performance of the \ocrt methods and their \rf implementations, we analyze each dataset in both noise-free and noisy conditions. Specifically, all output vectors strictly satisfy the underlying constraints in the noise-free datasets, whereas the noisy versions—generated by applying a Gaussian perturbation to the clean data— violate them.

The complete optimization model employed in HTS datasets is explicitly presented in Appendix~\ref{app:synthetic_optimization}.

\begin{table}[htbp]
  \centering
  \small
  {
    \caption{Results of HTS datasets for different tree depths with and without noise}
    \begin{tabular}{cclccr}
      \toprule
                                     & Depth                 & \ocrt Methods & $\Delta$ & MSE  & Time   \\
      \midrule
      \multirow{4}[4]{*}{Noise-free} & \multirow{2}[2]{*}{5} & \epocrt       & 45.46\%  & 7.17 & 0.93   \\
                                     &                       & \eocrt        & 42.71\%  & 7.03 & 71.04  \\
      \cmidrule{2-6}                 & \multirow{2}[2]{*}{7} & \epocrt       & 38.15\%  & 7.33 & 1.43   \\
                                     &                       & \eocrt        & 36.50\%  & 7.24 & 95.87  \\
      \midrule
      \multirow{4}[4]{*}{Noisy}      & \multirow{2}[2]{*}{5} & \epocrt       & 44.81\%  & 7.19 & 1.84   \\
                                     &                       & \eocrt        & 41.83\%  & 7.04 & 161.49 \\
      \cmidrule{2-6}                 & \multirow{2}[2]{*}{7} & \epocrt       & 38.28\%  & 7.36 & 2.24   \\
                                     &                       & \eocrt        & 36.20\%  & 7.24 & 174.33 \\
      \bottomrule
    \end{tabular}%
    \label{tab:hts_avg_depth}%
  }
\end{table}%

Table~\ref{tab:hts_avg_depth} reports the average percentage MSE gaps $\Delta$ relative to \skl, nominal MSE values, and the corresponding average training times (in seconds) for the hierarchical time‐series datasets.  We report the results for tree depths of 5 and 7. Both constraint‐aware methods (\epocrt and \eocrt) incur a positive MSE gap. Specifically, \eocrt achieves the smaller gap on both tree depths (on average equal to 39.31\%) compared to \epocrt (on average equal to 41.67\%), demonstrating that enforcing constraints during tree construction yields more accurate forecasts in the HTS setting. The results also show that deeper trees help capture more of the hierarchical structure, narrowing the gap to \skl. However, this accuracy gain comes at the expense of longer training times: \eocrt time increases from 116.26 seconds at depth 5 to 135.10 seconds at depth 7.  Interestingly, the results indicate that the relative performance of the \ocrt methods is closer to \skl on noisy datasets. Looking at the nominal MSE values, the introduction of noise increases the overall prediction error across all datasets. However, this deterioration is significantly more pronounced in the unconstrained \skl baseline than in the \ocrt methods. Finally, we observe that noisy datasets pose a greater computational challenge. This is mathematically intuitive: the underlying optimization problem is characteristically easier to solve when the training targets satisfy the constraints.

\section{Further Implications of Constrained Prediction with Decision Trees} \label{sec:extension}

Constrained prediction and our proposed methods can play a crucial role in many applications, particularly where traditional unconstrained prediction tools may fail to produce efficient or practically usable results. To illustrate such scenarios, we present two applications and some preliminary results that serve as representative examples of broader problem settings.  Additionally, we discuss a third application in Appendix~\ref{app:experiments}.

\subsection{Generalized Loss Functions} \label{sec:general_loss}

Traditional learning methods often rely on well-established loss functions such as mean squared error (MSE) or mean absolute deviation (MAD). These functions typically provide a good fit when the sole objective is to generate accurate predictions in a conventional sense. However, in many practical applications, it becomes necessary to consider alternative, and sometimes complex, loss functions that reflect different evaluation perspectives. In such cases, standard tools may fall short as they are often designed with conventional loss functions.

A similar challenge arises in the presence of latent variables, where the quantity of interest is not directly observable but depends on the target variables through a complex or non-trivial transformation. Regardless of whether the challenge stems from generalized loss functions or latent variables, such complications motivate the use of the proposed methods.

Let $g:\RR^K \mapsto \RR$ be a real-valued function. Suppose we aim to predict target values by minimizing a generalized loss function of the form $\ell(g(\vy),g(\hat{\vy}))$. Accordingly, the constrained prediction problem transforms into the following,
\[
  \vy_{\CS} = \arg \min_{\hat{\vy}\in \YY}\sum_{i \in \CS} \ell(g(\vy),g(\hat{\vy})).
\]

In this setting, conventional tools such as \skl cannot accommodate generalized loss functions. On the contrary, the approaches proposed herein are, by definition, able to accommodate such complexities. To illustrate this setting, we conducted a simple experimental study using previously generated synthetic datasets with $n=500$ and $K=5$, and set the maximum tree depth parameter to five. We define $g(\vy) = \sum_{i=1}^Kw_i\vy_i$ where $w_i$ represents a randomly drawn coefficient for target $i$. Hence, $g(\cdot)$ corresponds to a weighted sum of the target vector. We generated 3 randomly generated weight vectors.

\begin{table}[htbp]
  \centering
  \small
  \caption{Average MSE gaps with generalized loss function}
  \begin{tabular}{lr}
    \toprule
    Methods & \multicolumn{1}{c}{$\Delta_w$} \\
    \midrule
    \epocrt & -0.98\%                        \\
    \eocrt  & -8.54\%                        \\
    \bottomrule
  \end{tabular}%
  \label{tab:loss_avg}%
\end{table}%

Table~\ref{tab:loss_avg} presents the average loss, $\Delta_w$, of each method relative to the weighted loss value yielded by the standard \skl solution. The results demonstrate that both \eocrt and \epocrt methods provide better loss values as compared to \skl despite the fact that \skl's predictions are infeasible. Another observation is that the difference between \eocrt and \epocrt elevates as compared to our earlier experiments presented in the previous sections. This is an expected result as \epocrt constructs the tree with the conventional \skl approach and only considers the optimal constrained prediction problem at resulting leaf nodes. As such, the generalized loss function influences predictions only at the leaves and not during tree construction. This highlights the importance of incorporating the optimal prediction problem throughout the entire training process. Consequently, \eocrt\ and \mocrt\ emerge as the most promising alternatives for achieving high prediction accuracy in the presence of generalized loss functions.

\subsection{End-to-end Learning Framework}
Suppose we are given a dataset where the output vectors represent the objective function coefficients of a downstream optimization problem. The primary objective is to predict these target values for use in the downstream task. The traditional approach, known as predict-then-optimize, is a two-stage process: predictions are made first without knowledge of the downstream problem, and then simply plugged into the optimization model. Nevertheless, recent studies have demonstrated the effectiveness of considering the optimization and prediction problems jointly, which inherently accounts for the structure of the optimization problem. This approach is known as end-to-end learning or smart predict-then-optimize in the literature \citep{elmachtoub2022smart}. Consequently, it has become standard to minimize the suboptimality gap (or regret) of the downstream problem rather than directly minimizing prediction error. In this context, regret is defined as the difference between the objective value obtained from the predicted parameters and the true optimal objective value.

We formulate the downstream optimization problem as follows:
\[
  \min_{\vq \in \mathbb{R}^K}\{\vy^\top \vq : A\vq \geq \vb, \vq \geq 0 \},
\]
where $A \in \mathbb{R}^{m \times K}$ and $b \in \mathbb{R}^{m}$. Let $\vq_i$ denote the the true optimal solution for instance $i \in \CI$ which is obtained by solving the above problem after replacing $\vy$ with $\vy_i$.

Given a subset of sample indices $\mathcal{L}$ that denote a leaf of a tree, the prediction of the leaf $\hat{\vy}$ and the corresponding optimal decisions for the downstream optimization model $\hat{\vq}$ can be determined by solving the following bilevel problem:
\[
  \begin{array}{ll}
    \underset{\hat{\vy}, \hat{\vq}}{\text{minimize}} & \sum_{i \in \mathcal{L}} \| \vy_i^\top \vq_i - \vy_i^\top \hat{\vq}\|^2                           \\[2mm]
    \text{subject to }                               & \hat{\vq} \in \arg\min_{\vq \in \mathbb{R}^K}\{\hat{\vy}^\top \vq : A\vq \geq \vb, \vq \geq 0 \}, \\[2mm]
                                                     & \hat{\vy} \in \YY.
  \end{array}
\]
As the lower level model is a linear program, one can obtain a single-level bilinear model by using strong duality; see, for example, \citep{beck23bilevel}.


The end-to-end learning framework can be viewed as an extension of the generalized loss function application discussed previously. As with generalized loss functions, \skl is not well-equipped to respond to end-to-end learning frameworks where output constraints must be integrated. Using \skl as is directly refers to traditional predict-then-optimize approach. Therefore, we hereby evaluate the performance of the output constrained end-to-end learning compared to traditional predict-then-optimize approach where output constraints are not taken into account. To this end, we conducted a simple experimental study using previously generated synthetic datasets with $n=500$ and $K=5$ employing a continuous knapsack problem as the downstream model. The complete optimization model is presented in Appendix~\ref{app:synthetic_optimization}.

\begin{table}[htbp]
  \centering
  \small

  \caption{Average optimization regret gaps for end-to-end learning}
  \begin{tabular}{clcc}
    \toprule
    \multicolumn{1}{l}{Depth} & \ocrt Methods & \multicolumn{1}{c}{$\Delta_r$} & \multicolumn{1}{c}{$\Delta_{r2}$} \\
    \midrule
    \multirow{2}[2]{*}{5}     & \epocrt       & -13.37\%                       & -24.20\%                          \\
                              & \eocrt        & -26.00\%                       & -43.75\%                          \\
    \midrule
    \multirow{2}[2]{*}{7}     & \epocrt       & -11.05\%                       & -20.52\%                          \\
                              & \eocrt        & -22.03\%                       & -37.35\%                          \\
    \bottomrule
  \end{tabular}%
  \label{tab:regret}%
\end{table}%

Table~\ref{tab:regret} presents the regret gaps on test instances. Specifically, $\Delta_r$ denotes the average regret gap, whereas $\Delta_{r2}$ represents the average squared regret gap, both relative to corresponding regret values yielded by the baseline \skl solution. Although the end-to-end learning optimization model presented above explicitly minimizes squared regret, we report the standard regret gap ($\Delta_r$) to align with conventional literature. The results demonstrate that both \eocrt and \epocrt significantly reduce regret compared to \skl. This observation confirms the expectation that \skl would not be able to accommodate sophisticated end-to-end learning objectives. Furthermore, \eocrt outperforms \epocrt significantly, aligning with our previous findings on generalized loss functions and underscoring the importance of integrating the optimization problem directly into the training process.

\section{Conclusion.}
\label{sec:conc}

In this paper, we introduced new methods for incorporating output constraints directly into decision tree-based models while ensuring feasibility in multi-target regression tasks. These are called output-constrained-regression-tree (\ocrt) methods, and they differ in the approach they handle the feasibility: mixed-integer-programming (\mocrt), enumerative (\ocrt), and post-processing (\epocrt). By extending traditional decision trees with these constraint-aware learning approaches, we demonstrated that these methods enhance the practicality of predictions across a range of constrained settings. Our computational evaluations on synthetic datasets and hierarchical time series applications confirmed the efficacy of the proposed methods.

Our contribution is particularly valuable in areas where constrained feasibility is important, such as hierarchical forecasting, inventory management, logistics optimization, and healthcare resource allocation. These applications frequently require predictions that comply with physical, operational, or hierarchical constraints, which traditional unconstrained predictive models fail to satisfy. By integrating constraints directly into the tree-growing process, our methods avoid the inefficiencies and inaccuracies of post-hoc adjustments and deliver coherent predictions from the outset. Additionally, we extended these approaches to ensemble learning frameworks using random forest as an example, which highlights their broader applicability under convex feasible sets.

There are, however, a few limitations to the proposed approaches. First, while \mocrt provides flexibility for producing feasible predictions, its reliance on mixed-integer programming introduces computational challenges, especially for larger datasets or deeper trees. Similarly, while \eocrt offers a computational advantage over \mocrt, its evaluation of all possible splits becomes challenging when the number of features or split thresholds is large. The simplified \epocrt method, despite being computationally efficient, sacrifices accuracy by addressing constraints only at the leaf nodes. Furthermore, the feasibility guarantees of our proposed ensemble methods assume convex feasible sets, limiting their applicability to more complex or non-convex constraints.


\clearpage

\clearpage
\bibliographystyle{elsarticle-harv}
\bibliography{ConPred}


\clearpage

\appendix

\section{}
\label{prf:theorem}
Here, we provide a proof for Theorem \ref{thm1}.

\textsc{Theorem \ref{thm1}:} \textit{
  Let $\YY \subseteq \RR^k$ be convex and $y_i\in \YY$ for all $i\in \CI$. If the loss function is the mean squared error or the mean Poisson deviance, then the mean of the target vectors in a node minimizes the loss.}

\begin{proof} Suppose that the node corresponds to the subset  $\CN \subseteq \CD$. Since $\YY \subseteq \RR^K$, we have
  \[
    \min_{\hat{\vy} \in \RR^K} ~ \sum_{i \in \CI_{\CN}}\ell(\hat{\vy}, \vy_i) \leq \min_{\hat{\vy} \in \YY} ~ \sum_{i \in \CI_{\CN}}\ell(\hat{\vy}, \vy_i)
  \]
  for any loss function $\ell:\RR^K \times \RR^K \mapsto \RR$. We call the optimization model on the left as \textit{the unconstrained problem} and the one on the right as \textit{the constrained problem}. Note that if the optimal solution of the unconstrained problem is in the feasible set $\YY$, then it should also be the optimal solution of the constrained problem.  We use the optimal solution of the unconstrained problem as the output vector of the node and refer to it as $\vy_\CN$. Both loss functions that we consider are separable in terms of the target vector. Thus, we write $\ell(\vy, \vy_i) = \sum_{t\in\CT} \ell(y_t, y_{it})$, where $\CT = \{1, \dots, K\}$ is the set of target indices.

  Below, we first focus on the first part and then discuss the second part of the theorem. The unconstrained problems with the mean squared error and the mean Poisson deviance loss functions are given by
  \[
    \min_{\hat{\vy} \in \RR^k} ~ \frac{1}{2|\CI_{\CN}|}\sum_{i \in \CI_{\CN}} \sum_{t \in \CT} (\hat{y}_t - y_{it})^2
  \]
  and
  \[
    \min_{\hat{\vy} \in \RR^k} ~ \frac{1}{|\CI_{\CN}|}\sum_{i \in \CI_{\CN}} \sum_{t \in \CT} (y_{it} \ln(\frac{y_{it}}{\hat{y}_t}) + \hat{y}_t - y_{it}),
  \]

  respectively. Clearly, both loss functions are differentiable and convex. Thus, the necessary-and-sufficient optimality conditions for both problems are given by
  \begin{equation}
    \label{eq:fop}
    \sum_{i \in \CI_{\CN}} \frac{\partial \ell(\hat{y}_t, y_{it})}{\partial \hat{y}_t} = 0, \ \ t \in \CT.
  \end{equation}

  Solving next the conditions in \eqref{eq:fop} for both unconstrained problems leads to
  \[
    \sum_{i \in \CI_{\CN}} \frac{\partial\ell(\hat{y}_t, y_{it})}{\partial \hat{y}_t} = 0, \ t \in \CT \implies
    \hat{y}_t = \frac{1}{|\CI_{\CN}|} \sum_{i \in \CI_{\CN}} y_{it}, \ t \in \CT \implies \vy_{\CN} = \frac{1}{|\CI_{\CN}|} \sum_{i \in \CI_{\CN}} \vy_i.
  \]
  Thus, the optimal solutions is the mean vector. Because $\vy_i \in \YY$ for $i \in \CI_{\CN}$ and the feasible set $\YY$ is convex, the convex combination of $\vy_i$ vectors should also be feasible; that is, $\vy_\CN \in \YY$. Therefore, the mean vector is also the optimal solution of the constrained problem. This shows the desired result for the first part.
\end{proof}

\begin{remark} \label{rem:MAE}
  In unconstrained multi-target regression under mean absolute error loss, the median target vector in a node minimizes the loss. However, even when the feasible set $\YY \subseteq \RR^k$ is convex and $y_i\in \YY$ for all $i\in \CI$, the median vector may lie outside the feasible set. To illustrate this, consider the unit vectors in three-dimensional space. The median vector will be the origin, but it does not lie in the simplex defined by the unit vectors.
\end{remark}

\vspace{0.5cm}

\clearpage

\section{} \label{app:synthetic_optimization} Below, we provide the optimization models that we used in our computational study.

\noindent\textbf{Synthetic Datasets. } The optimization model imposed on \textit{synthetic} datasets is formulated as follows:
\[
  \begin{array}{lll}
    \minimize & \frac{1}{2t}\sum_{i=1}^{t} \sum_{k=1}^{K} (\hat{y}_k - y_{ik})^2 &                                                         \\[2mm]
    \subto    & \hat{y}_{k_j-1}-\hat{y}_{k_j} = \frac{j+1}{10},                  & j\in  \left[0,\floor{\frac{K-\floor{K/2}}{2}}-1\right], \\[2mm]
              & \sum_{k=1}^{\floor{K/2}}\hat{y}_{k} = 1,                         &                                                         \\[2mm]
              & \hat{y}_k\geq 0,                                                 & k\in [1,K],
  \end{array}
\]
where $k_j=\floor{\frac{K}{2}}+2j$, \( t \) is the number of instances, \( K \) is the number of targets, \( \hat{y}_t \) is the prediction for the \( k \)-th target, and \( y_{ik} \) is the true value of the \( k \)-th target for the \( i \)-th instance.

\noindent\textbf{Hierarchical Time Series Datasets. }
The optimization model imposed on \textit{hts} datasets is formulated as follows:
\[
  \begin{array}{lll}
    \minimize & \frac{1}{2t}\sum_{i=1}^{t} \sum_{k=1}^{K} (\hat{y}_k - y_{ik})^2 &                         \\[2mm]
    \subto    & \hat{y}_{k} \leq M z_k,                                          & k\in  \left[1,K\right], \\[2mm]
              & \sum_{k=1}^{K}\hat{y}_{k} = 15,                                  &                         \\[2mm]
              & \sum_{k=1}^{K}z_{k} \leq 4,                                      &                         \\[2mm]
              & \hat{y}_k\geq 0, z_k\in \{0,1\},                                 & k\in [1,K],
  \end{array}
\]
where \( t \) is the number of instances, \( K \) is the number of targets, \( M \) is a sufficiently large number, \( \hat{y}_t \) is the prediction for the \( k \)-th target, and \( y_{ik} \) is the true value of the \( k \)-th target for the \( i \)-th instance.

\noindent\textbf{End-to-end learning framework. }
The optimization problem, a variant of the knapsack problem integrated into the end-to-end learning framework, is formulated as follows:
\[
  \begin{array}{lll}
    \maximize & \sum_{k=1}^{K} \hat{y}_kq_k              &             \\[2mm]
    \subto    & \sum_{k=1}^{\floor{K/2}}q_{k} \leq 100,  &             \\[2mm]
              & \sum_{k=\floor{K/2}+1}^{K}q_{k} \leq 10, &             \\[2mm]
              & q_k\geq 0,                               & k\in [1,K],
  \end{array}
\]
where $\hat{y}_k$ denotes the prediction for the \( k \)-th target, i.e. the profit of item $k$, and $q_k$ represents the amount of item $k$ put into the knapsack. The integrated model, designed to jointly determine optimal predictions and decision variables to minimize regret, now becomes:
\[
  \begin{array}{lll}
    \minimize & \sum_{i=1}^{t} \left(\sum_{k=1}^{K} y_{ik}q_{ik} - y_{ik}\hat{q}_k\right)^2 &                                                         \\[2mm]
    \subto    & \sum_{k=1}^{K} \hat{y}_k \hat{q}_k \geq 100\pi_1 + 10\pi_2 ,                &                                                         \\[2mm]
              & \sum_{k=1}^{\floor{K/2}}\hat{q}_{k} \leq 100,                               &                                                         \\[2mm]
              & \sum_{k=\floor{K/2}+1}^{K}\hat{q}_{k} \leq 10,                              &                                                         \\[2mm]
              & \hat{y}_{k} \leq \pi_1,                                                     & k\in [1,\floor{K/2}]                                    \\[2mm]
              & \hat{y}_{k} \leq \pi_2,                                                     & k\in [\floor{K/2}+1,K]                                  \\[2mm]
              & \hat{y}_{k_j-1}-\hat{y}_{k_j} = \frac{j+1}{10},                             & j\in  \left[0,\floor{\frac{K-\floor{K/2}}{2}}-1\right], \\[2mm]
              & \sum_{k=1}^{\floor{K/2}}\hat{y}_{k} = 1,                                    &                                                         \\[2mm]
              & \hat{y}_k, \hat{q}_k\geq 0,                                                 & k\in [1,K],                                             \\[2mm]
              & \pi_1, \pi_2\geq 0,
  \end{array}
\]
where $\pi_i$ denotes the dual variable associated with constraint $i$ in the knapsack problem, $q_{ik}$ represents the optimal quantity of item $k$ for instance $i$. Further, $\hat{y}_k$ denotes the prediction for the $k$-th item, whereas $\hat{q}_k$ represents the quantity of item $k$ selected for the corresponding leaf node.

\clearpage

\section{}
\label{app:experiments}
\noindent\textbf{Further Tests on Synthetic Datasets.}
We conduct additional experiments to evaluate the performance of the proposed methods on larger datasets. For this analysis, we employ a focused experimental design. We define a base instance using a synthetic dataset with $n=1000$ observations and $p=6$ features. We then generate new instances by varying one parameter at a time, testing three distinct values for each: $n \in \{1000, 5000, 10000\}$ and $p \in \{6, 9, 12\}$. For every parameter configuration, we randomly generate three replicate datasets following the procedure detailed in the previous section. Finally, because our earlier results clearly demonstrated that the $\mocrt$ method does not scale well with an increasing number of instances, we have excluded it from the scalability tests on the number of instances.

\begin{table}[htbp]
  \centering
  \small
  {
    \caption{Average MSE gaps on synthetic datasets for different numbers of features}
    \begin{tabular}{clr|rr}
      \toprule
      \multicolumn{1}{c}{$p$} & \ocrt Methods & \multicolumn{1}{c|}{$\Delta$} & \multicolumn{1}{l}{\rf Methods} & \multicolumn{1}{c}{$\Delta$} \\
      \midrule
      \multirow{3}[2]{*}{6}   & \epocrt       & 53.58\%                       & \multicolumn{1}{l}{\prf}        & 54.92\%                      \\
                              & \eocrt        & 52.21\%                       & \multicolumn{1}{l}{\erf}        & 52.30\%                      \\
                              & \mocrt        & 52.34\%                       &                                 &                              \\
      \midrule
      \multirow{3}[2]{*}{9}   & \epocrt       & 19.44\%                       & \multicolumn{1}{l}{\prf}        & 6.43\%                       \\
                              & \eocrt        & 12.21\%                       & \multicolumn{1}{l}{\erf}        & -0.60\%                      \\
                              & \mocrt        & 12.92\%                       &                                 &                              \\
      \midrule
      \multirow{3}[2]{*}{12}  & \epocrt       & 27.47\%                       & \multicolumn{1}{l}{\prf}        & 16.33\%                      \\
                              & \eocrt        & 25.43\%                       & \multicolumn{1}{l}{\erf}        & 13.29\%                      \\
                              & \mocrt        & 25.22\%                       &                                 &                              \\
      \bottomrule
    \end{tabular}%
    \label{tab:further_feature}%
  }
\end{table}%

\begin{table}[htbp]
  \centering
  \small
  {
    \caption{Average MSE gaps on synthetic datasets for larger number of instances}
    \begin{tabular}{clr|lr}
      \toprule
      $K$                       & \ocrt Methods & \multicolumn{1}{c|}{$\Delta$} & \rf Methods & \multicolumn{1}{c}{$\Delta$} \\
      \midrule
      \multirow{2}[2]{*}{1000}  & \epocrt       & 53.58\%                       & \prf        & 54.92\%                      \\
                                & \eocrt        & 52.21\%                       & \erf        & 52.30\%                      \\
      \midrule
      \multirow{2}[2]{*}{5000}  & \epocrt       & 59.96\%                       & \prf        & 63.23\%                      \\
                                & \eocrt        & 58.78\%                       & \erf        & 60.66\%                      \\
      \midrule
      \multirow{2}[2]{*}{10000} & \epocrt       & 60.77\%                       & \prf        & 64.00\%                      \\
                                & \eocrt        & 59.14\%                       & \erf        & 60.71\%                      \\
      \bottomrule
    \end{tabular}
    \label{tab:further_instance}%
  }
\end{table}%


Tables \ref{tab:further_feature} and \ref{tab:further_instance} report the average MSE gaps. The results remain consistent with our earlier findings: \eocrt stands out as the best alternative among the \ocrt methods, whereas \erf emerges as the superior choice among the \rf methods. A notable observation from these tables is that the predictive performance of the \ocrt methods becomes much closer to the unconstrained \skl baseline in datasets with a higher number of features. This probably occurs because the functional relationship between the input and output spaces shifts as the feature dimension varies. Consequently, this demonstrates that there are likely specific functional relationships where constrained \ocrt methods (or their \rf counterparts) can actually outperform \skl, despite the latter producing infeasible predictions. A similar trend is evident in the performance gap between \eocrt and \epocrt. Depending heavily on the underlying functional mapping between the input and output spaces, the predictive advantage of \eocrt over \epocrt can become significantly pronounced. Beyond these insights, the remaining results align with our previously discussed observations.

\begin{figure}[htbp]
  \centering
  \begin{subcaptionbox}{Average training time per method (log scale) by size\label{fig2:fig1}}[0.48\textwidth]
    {\includegraphics[width=\linewidth]{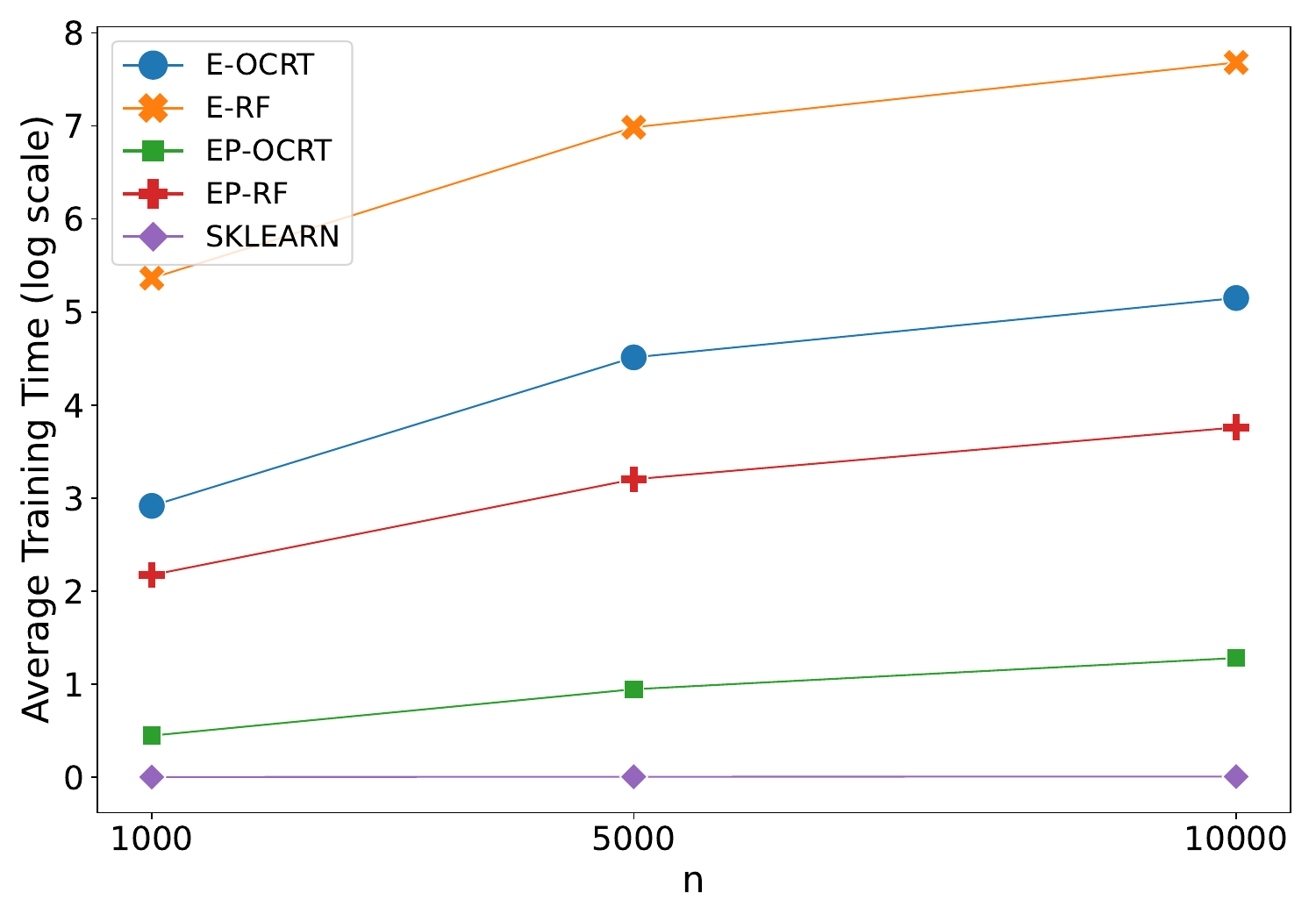}}
  \end{subcaptionbox}
  \hspace{0.009\textwidth}
  \begin{subcaptionbox}{Average training time (log scale) per method by number of features \label{fig2:fig2}}[0.48\textwidth]
    {\includegraphics[width=\linewidth]{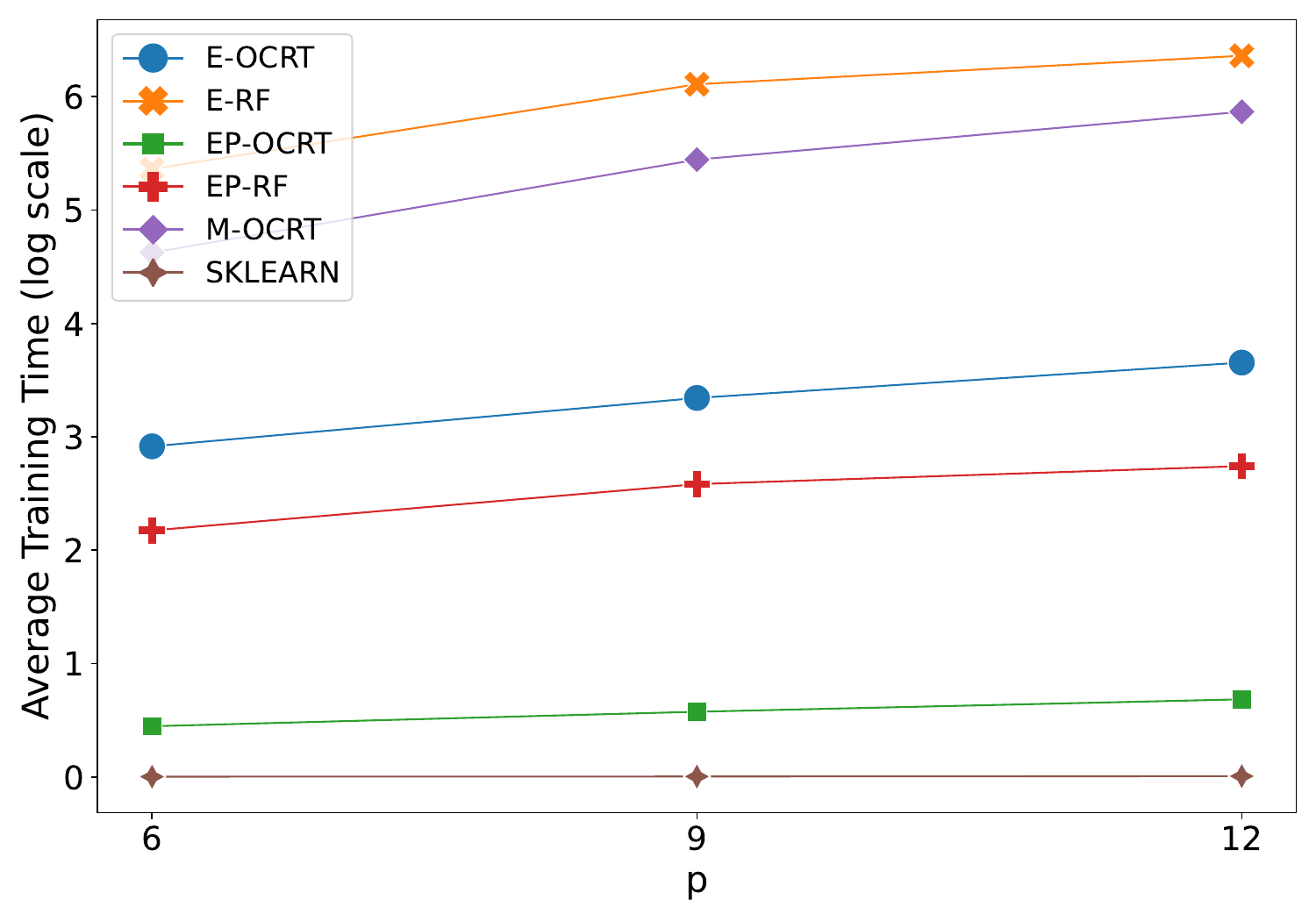}}
  \end{subcaptionbox}
  \caption{Average training times}
  \label{fig:ttime2}
\end{figure}

Figure~\ref{fig:ttime2} illustrates the average computation times for each method across the additional experiments. Specifically, the left panel (Figure~\ref{fig2:fig1}) demonstrates how computation time scales with dataset size, whereas the right panel (Figure~\ref{fig2:fig2}) displays scalability with respect to feature dimensionality.  These results remain entirely consistent with our earlier findings. Importantly, they highlight that both the \eocrt and \erf methods can efficiently process larger datasets within reasonable computation times.

\noindent\textbf{Optimization with Constraint Learning.}
In an OCL framework, predictive models are embedded within optimization problems to support decision-making.  It is important to clarify that in this case, contrary to the end-to-end framework, the predictive model is already trained, meaning that the parameters of the predictive model are fixed, however, and the features are also variables of the optimization model:
\[
  \min\{f(\vx, \hat{\vy})~:~\hat{\vy} = \hat{h}(\vx), \hat{\vy} \in \YY, \vx \in \mathbb{X}\},
\]
where $\vx$ is the decision variable of the optimization model but it is also the (sub-)set of features used in the fitted model $\hat{h}(\cdot)$. Here, $\vy$ is the outcome of the predictive model and it can be further constrained or/and used in the objective function.

To illustrate the importance of \ocrt, we use an example from logistics, where an operations manager must determine how to distribute products from a central warehouse to various retail stores in a way that maximizes profit. The quantities to be shipped, indicated with $\hat{\vy}$ in the above model, are predicted using a machine learning model trained on historical data that considers variables such as product price, seasonality, and marketing spend, corresponding to the vector $\vx$. However, the optimization problem must also satisfy real-world constraints, such as storage limitations at each store (indicated with $\mathbb{Y}$), a cap on the number of stores receiving each product, and a global marketing budget (indicated with $\mathbb{X}$).

If the predictive model produces shipment quantities that exceed these limits, the resulting optimization problem becomes infeasible, and no valid distribution plan exists that respects all the constraints. In such situations, decision-makers are left with predictions that are not actionable and must be manually adjusted, at the cost of optimality and efficiency. Having incoherent outputs that do not satisfy downstream constraints makes the decision-making problem infeasible.
To address this issue, decision-makers must either manually adjust the solutions, which can be particularly challenging in problems involving multiple decision variables and constraints, or reformulate the optimization model to allow controlled violations via slack variables, penalizing their magnitude in the objective function. While the introduction of slack variables may seem like a practical way to handle predictive outputs that do not inherently satisfy the constraints, this approach effectively amounts to post-processing. If the objective consists solely of minimizing the slack variables, the resulting formulation is equivalent to \epocrt. However, when the objective includes additional terms, the minimization of slack variables may conflict with these competing objectives. In such cases, the formulation is no longer equivalent to \epocrt and may even move further away from the original goal: obtaining a solution that both satisfies the constraints and remains as close as possible to the prediction. As demonstrated in Section~5 and Section~6.2, these adjustments can yield suboptimal solutions that deviate more significantly from the ground truth, ultimately reducing overall predictive accuracy.

This highlights the importance of using output-constrained predictive models like \ocrt. These models can be trained to respect key structural or numerical constraints during prediction, ensuring that the optimization model has feasible solutions. Moreover, decision trees are widely adopted in OCL given their efficient embedding into optimization problems, and their interpretable structure, which allows for higher transparency in high-stakes applications \citep{Verwer2017, Maragno2024Embedding}.

Beyond logistics, this issue arises in several other domains such as:
\begin{itemize}
  \item Power grid management: predictive models are used to estimate electricity loads across substations. If the predicted loads violate network or capacity constraints, the resulting power flow problem may be unsolvable, risking grid instability.
  \item Healthcare resource allocation: predicting the number of patients requiring different types of care must align with staffing and equipment constraints. Violations in these predictions can lead to infeasible or unsafe resource planning.
\end{itemize}

In all these cases, if constraints are not respected by the predictive model, the entire decision-support pipeline can break down. Thus, ensuring that machine learning outputs are consistent with domain constraints is essential.

\end{document}